\newif\ifconfver
\confverfalse      
\confvertrue        

\newif\ifonecoltab

\newif\ifplainver  

\ifplainver
    \confverfalse                         
\fi

\ifconfver
     \documentclass[10pt,twocolumn,twoside]{IEEEtran}
\else
    \ifplainver
        \documentclass[11pt]{article}
        \usepackage{fullpage}
    \else
        \documentclass[12pt,draftcls,onecolumn]{IEEEtran}
    \fi
\fi

\usepackage{calc,amsfonts,amssymb,amsmath,bm,url,color,theorem,graphicx,cite, dsfont}
\usepackage{psfrag,subfigure,float}
\usepackage[algoruled,linesnumbered]{algorithm2e}
\usepackage{multirow}
\usepackage{epstopdf}

\definecolor{orange}{RGB}{255,107,0}

\newtheorem{Property}{Property}

\theorembodyfont{\rmfamily}

\newcommand{\X}{\mathbf{X}}

\newcommand{\W}{\mathbf{W}}

\renewcommand{\H}{\mathbf{H}}
\newcommand{\D}{\mathbf{D}}
\newcommand{\A}{\mathbf{A}}
\newcommand{\B}{\mathbf{B}}
\newcommand{\C}{\mathbf{C}}
\newcommand{\I}{\mathbf{I}}

\newcommand{\Z}{\mathbf{Z}}
\newcommand{\U}{\mathbf{U}}
\newcommand{\V}{\mathbf{V}}

\renewcommand{\S}{\mathbf{S}}

\newcommand{\M}{\mathbf{M}}

\newcommand{\R}{\mathbb{R}}

\newcommand{\Xt}{\underline{\X}}

\newcommand{\etal}{\textit{et al. }}
\begin{document}

\bibliographystyle{IEEEtran}

\newcommand{\papertitle}{
Learning From Hidden Traits: Joint Factor Analysis and Latent Clustering
}


\newcommand{\paperabstract}{
Dimensionality reduction techniques play an essential role in data analytics, signal processing and machine learning. Dimensionality reduction is usually performed in a preprocessing stage that is separate from subsequent data analysis, such as clustering or classification. Finding reduced-dimension representations that are well-suited for the intended task is more appealing. This paper proposes a joint factor analysis and latent clustering framework, which aims at learning cluster-aware low-dimensional representations of matrix and tensor data. The proposed approach leverages matrix and tensor factorization models that produce essentially unique latent representations of the data to unravel latent cluster structure -- which is otherwise obscured because of the freedom to apply an oblique transformation in latent space. At the same time, latent cluster structure is used as prior information to enhance the performance of factorization. Specific contributions include several custom-built problem formulations, corresponding algorithms, and discussion of associated convergence properties. Besides extensive simulations, real-world datasets such as Reuters document data and MNIST image data are also employed to showcase the effectiveness of the proposed approaches.
}


\ifplainver

    \date{\today}

    \title{\papertitle\footnote{Part of this work was published in {\em Proc. IEEE CAMSAP 2015} \cite{yangjoint}.}}

    \author{
     $^\ast$Bo Yang,  $^\ast$Xiao Fu,  and $^\ast$Nicholas, D. Sidiropoulos
    \\ ~ \\
		$^\ast$Dept Elec. Computer Eng., University of Minnesota,\\
		Minneapolis, 55455, MN, United States\\
		Email: (yang4173,xfu,nikos)@umn.edu
    }

    \maketitle

    \begin{abstract}
    \paperabstract
    \end{abstract}

\else
    \title{\papertitle}

    \ifconfver \else {\linespread{1.1} \rm \fi

   \author{Bo Yang, \IEEEmembership{Student Member, IEEE}, Xiao Fu, \IEEEmembership{Member, IEEE}, Nicholas D. Sidiropoulos$^\dag$, \IEEEmembership{Fellow, IEEE}
\thanks{Preliminary version of part of this work appears in {\em Proc. IEEE CAMSAP 2015} \cite{yang2016joint}.}
\thanks{ B. Yang, X. Fu and N.D. Sidiropoulos are with the Department of Electrical and Computer Engineering, University of Minnesota, Minneapolis, MN55455, e-mail (yang4173,xfu,nikos)@umn.edu. Supported in part by NSF IIS-1447788, IIS-1247632.}
}

    \maketitle

    \ifconfver \else
        \begin{center} \vspace*{-2\baselineskip}
        \end{center}
    \fi

    \begin{abstract}
    \paperabstract
    \end{abstract}


    \ifconfver \else \IEEEpeerreviewmaketitle} \fi

 \fi

\ifconfver \else
    \ifplainver \else
        \newpage
\fi \fi

\section{Introduction}
Many signal processing and machine learning applications nowadays involve high-dimensional raw data that call for appropriate compaction before any further processing. Dimensionality reduction (DR) is often applied before clustering and classification, for example. Matrix and tensor factorization (or {\em factor analysis}) plays an important role in DR of matrix and tensor data, respectively. Traditional factorization models, such as singular value decomposition (SVD) and principal component analysis (PCA) have proven to be successful in analyzing high-dimensional data -- e.g., PCA has been used for noise suppression, feature extraction, and subspace estimation in numerous applications. In recent years, alternative models such as non-negative matrix factorization (NMF) \cite{lee1999learning, gillis2014and} have drawn considerable interest (also as DR tools), because they tend to produce unique and interpretable reduced-dimension representations. In parallel, tensor factorization for multi-way data continues to gain popularity in the machine learning community, e.g., for social network mining and latent variable modeling \cite{maruhashi2011multiaspectforensics,agrawal2015study,papalexakis2015location,davidson2013network,jeon2015haten2}.

\begin{figure}
	\centering
		\includegraphics[width=8cm]{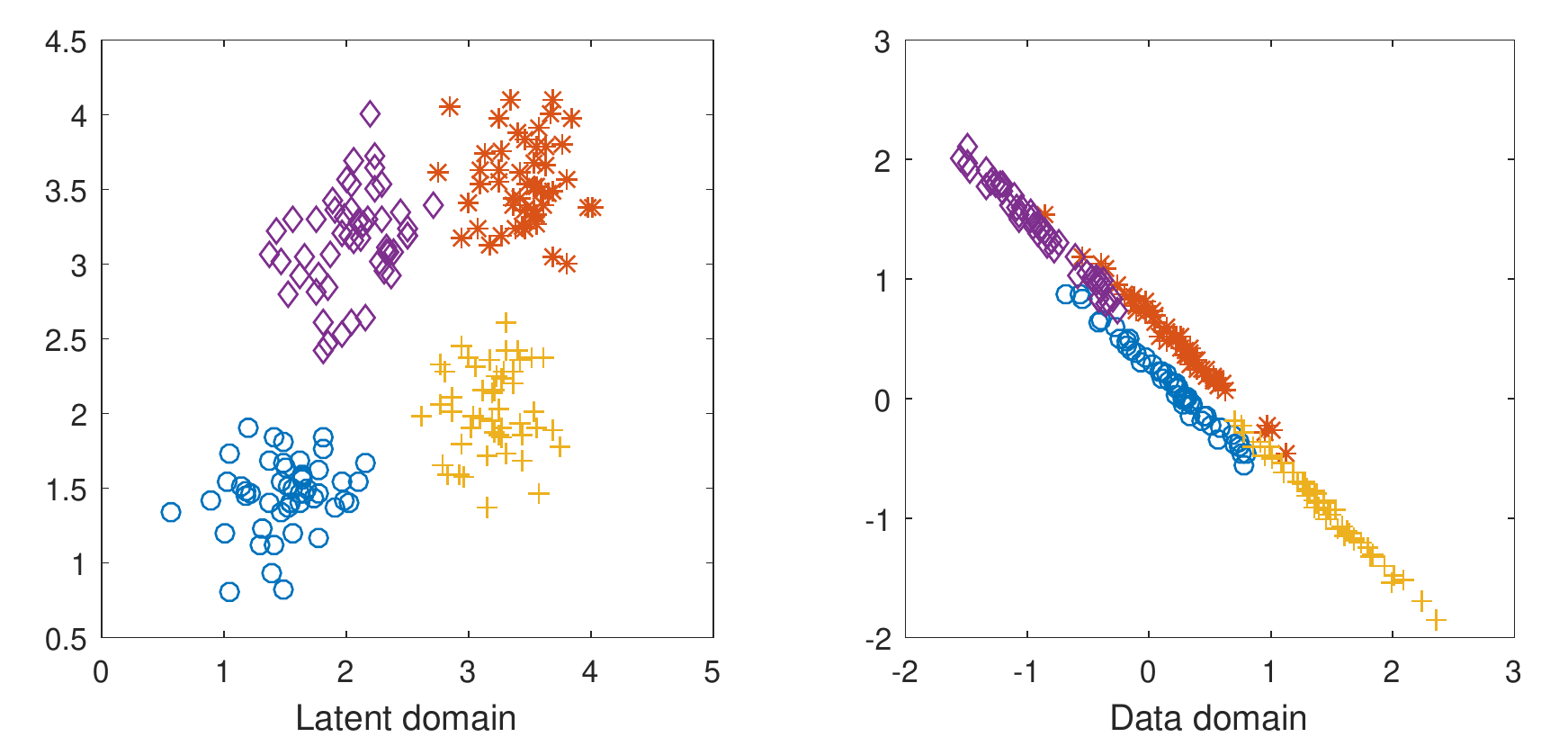}
	\caption{The effect of distance distortion introduced by ${\bf W}$. Left: ${\bf H}$ on a 2-dimensional plane. Right: ${\bf X}={\bf W}{\bf H}$ on a 2-dimensional plane.}
	\label{fig:motivation_W}
\end{figure}

When performing DR or factor analysis, several critical questions frequently arise.
First, what type of factor analysis should be considered for producing useful data representations for further processing, e.g., classification and clustering? Intuitively, if the data indeed coalesce in clusters in {\em some} low-dimensional representation, then DR should ideally map the input vectors to this particular representation -- identifying the right subspace is not enough, for linear transformation can distort cluster structure (cf. Fig. \ref{fig:motivation_W}). Therefore, if the data follow a factor analysis model that is unique (e.g., NMF is unique under certain conditions \cite{donoho2003does,huang2014non}) {\em and} the data form clusters in the latent domain, then fitting that factor analysis model will reveal those clusters.

The second question is what kind of prior information can help get better latent representations of data?
Using prior information is instrumental for fending against noise and modeling errors in practice, and thus is well-motivated.
To this end, various constraints and regularization priors have been considered for matrix and tensor factorization, e.g., sparsity, smoothness, unimodality, total variation, and nonnegativity \cite{papalexakis2013k,chemometrics1998least,zhang2008total}, to name a few.

\begin{figure}
	\centering
		\includegraphics[width=5cm]{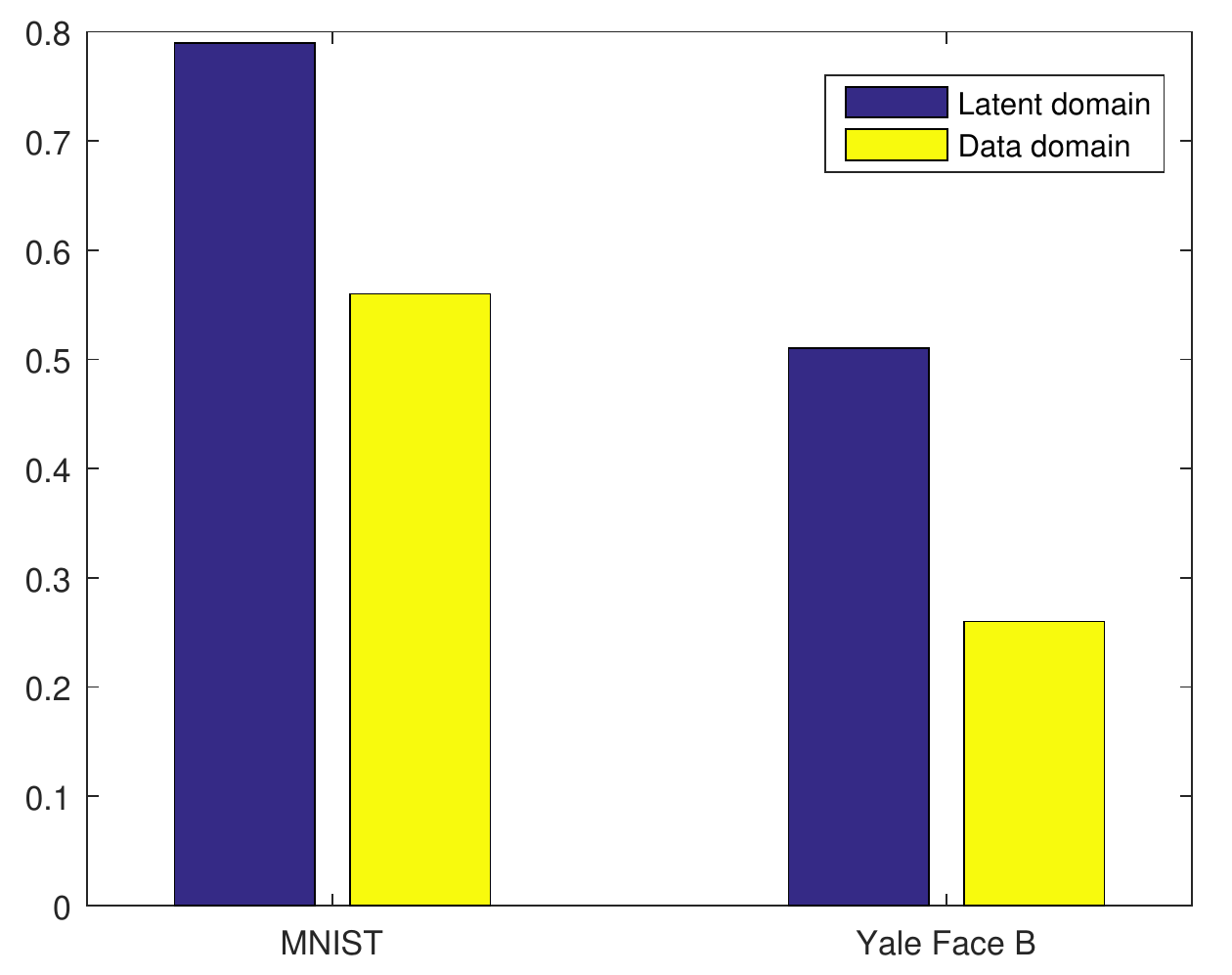}
	\caption{Contrasting data-domain and latent domain distance: The average cosine distance between 2500 pairs of data samples from different clusters. Data are taken from MNIST handwritten digits dataset and Yale Face B dataset, and the latent representations are found via NMF \cite{lee1999learning}.}
	\label{fig:motivation}
\end{figure}

In this work, we consider using a new type of prior information to assist factor analysis, namely,
the latent cluster structure.
This is motivated by the fact the dimension-reduced data
usually yields better performance in clustering tasks, which suggests that the cluster structure of the data is more pronounced in {\em some} latent domain relative to the data domain.
Some evidence can be seen in Fig.~\ref{fig:motivation}, where we compare the average data-domain and latent domain cosine distances\footnote{The cosine distance between two vectors ${\bf x}$ and ${\bf y}$ is defined as ${\rm d}({\bf x},{\bf y})=1-{\bf x}^T{\bf y}/(\|{\bf x}\|_2\|{\bf y}\|_2)$.}
of data points from two clusters of image data from the Yale B \footnote{Online available: http://web.mit.edu/emeyers/www/face\_databases.html} face image database and the MNIST handwritten digit image database\footnote{Online available: http://yann.lecun.com/exdb/mnist/}, where the latent representations are produced via NMF. We see that the average latent domain distance between the data from two different clusters is significantly larger than the corresponding data domain distance in both cases.
This observation motivates us to make use of such a property to enhance the performance of both factor analysis and clustering.

Using clustering to aid NMF-based DR was considered in \cite{cai2011locally,cai2011graph}, where a distance graph of the data points was constructed and used as
regularization for NMF -- which essentially forces reduced-dimension representations to be far (close) in the latent domain, if the high-dimension vectors are far (resp. close) in the data domain. However, the data domain distance and the latent domain distance are not necessarily proportional to each other, as is evident from Fig. \ref{fig:motivation}.

To see this clearly, consider a matrix factorization model ${\bf X}={\bf W}{\bf H}$ where each column of ${\bf X}$ represents a data point,
and the corresponding column of ${\bf H}$ is its latent representation.
Consider the squared distance between the first two columns of ${\bf X}$, i.e., $||{\bf X}(:,1)-{\bf X}(:,2)||_2^2 =({\bf H}(:,1)-{\bf H}(:,2))^T {\bf W}^T {\bf W} ({\bf H}(:,1)-{\bf H}(:,2))$,
where $:$ stands for all values of the respective argument. On the other hand, the distance of the latent representations of the first two columns of ${\bf X}$ is given by
$\left\| {\bf H}(:,1)-{\bf H}(:,2)\right\|_2^2
= ({\bf H}(:,1)-{\bf H}(:,2))^T({\bf H}(:,1)-{\bf H}(:,2))$.
Notice how the matrix ${\bf W}^T {\bf W}$ weighs the latent domain distance to produce the data domain distance; also see Fig.~\ref{fig:motivation_W} for illustration.

An earlier line of work \cite{de1994k,vichi2001factorial} that is not as well-known in the signal processing community considered latent domain clustering on the factor ${\bf H}$, while taking the other factor (${\bf W}$) to be a semi-orthogonal matrix.
However this cannot mitigate the weighting effect brought by ${\bf W}^T {\bf W}$, since orthogonal projection
cannot cancel the distorting effect brought in by the left factor ${\bf W}$.

Aiming to make full use of the latent cluster structure, we propose a novel {\em joint} factor analysis and latent clustering framework in this paper. Our goal is to identify the \emph{unique} latent representation of the data in some domain which is discriminative for clustering purposes, and also to use the latent cluster structure to help produce more accurate factorization results at the same time. We propose several pertinent problem formulations to exemplify the generality and flexibility of the proposed framework, and devise corresponding computational algorithms that build upon alternating optimization, i.e., alternating between factorization and clustering, until convergence. Identifiability of the latent factors plays an essential role in our approach, as it counteracts the distance distortion effect illustrated in Fig.~\ref{fig:motivation_W}. This is a key difference with relevant prior work such as \cite{de1994k,vichi2001factorial,patel2013latent}.

We begin with $K$-means latent clustering using several identifiable factorization models of matrix and tensor data, namely, nonnegative matrix factorization, convex geometry (CG)-based matrix factorization model \cite{fu2015blind}, and low-rank tensor factorization or {\em parallel factor analysis} (PARAFAC) \cite{harshman1970foundations}.
Carefully designed optimization criteria are proposed, detailed algorithms are derived, and convergence properties are discussed. We next consider extension to joint factorization and subspace clustering, which is motivated by the popularity of subspace clustering \cite{vidal2010tutorial}. The proposed algorithms are carefully examined in judiciously designed simulations.
Real experiments using document, handwritten digit, and three-way E-mail datasets are also used to showcase the effectiveness of the proposed approach.

\emph{Notation:}
Capital letters with underscore denote tensors, e.g. $\Xt$; bold capital letters $\A, \B, \C$ denote matrices; $\odot$ denotes the Khatri-Rao (column-wise Kronecker) product; $k_{\A}$ denotes the Kruskal rank of matrix $\A$; $\A^T$ denotes the transpose of $\A$ and $\A^\dagger$ denotes the pseudo inverse of $\A$; $\A(i,:)$ and $\A(:, j)$ denote the $i$th row and the $j$th column of $\A$, respectively; $\Xt(:,:,k)$ denotes the $k$th matrix slab of the three-way tensor $\Xt$ taken perpendicular to the third mode; and likewise for slabs taken perpendicular to the second and first mode, $\Xt(:,j,:)$, $\Xt(i,:,:)$, respectively; $\|{\bf x}\|_0$ counts the non-zero elements in the vector ${\bf x}$; calligraphic letters denote sets, such as  ${\cal J}$, ${\cal F}$. $\left\|\cdot \right\|_F$, $\left\|\cdot \right\|_2$, $\left\|\cdot \right\|_1 $ denote the Frobenius norm, $\ell_2$-norm and $\ell_1$-norm, respectively. ${\cal R}(\A)$ denotes the range space of matrix $\A$.

\section{Background}
In this section, we briefly review the pertinent prior art in latent clustering and factor analysis.
\subsection{Clustering and Latent Clustering}
Let us begin with the classical $K$-means problem \cite{macqueen1967some}:
Given a data matrix ${\bf X}\in\mathbb{R}^{I\times J}$, we wish to group the columns of ${\bf X}$ into $K$ clusters;
i.e., we wish to assign the column index of ${\bf X}(:,j)$ to cluster
${\cal J}_k$, $k \in \left\{1,\ldots,K\right\}$, such that ${\cal J}_1\cap \cdots \cap{\cal J}_K=\emptyset$, ${\cal J}_1\cup \cdots \cup{\cal J}_K=\{1,\ldots,J\}$, and the sum of intra-cluster dispersions is minimized.
The $K$-means problem can be written in optimization form as follows:
\begin{equation}\label{eq:km}
\begin{aligned}
             \min_{{\bf S}\in\mathbb{Z}^{K\times J},{\bf M}\in\mathbb{R}^{I\times K}}&\quad\|{\bf X}-{\bf M}{\bf S}\|_F^2\\
						                      {\rm s.t.}&\quad\|{\bf S}(:,j)\|_0=1,~{\bf S}(i,j)\in\{0,1\},\\
\end{aligned}
\end{equation}
where the matrix ${\bf S}$ is an assignment matrix,
${\bf S}(k,j)=1$ means that ${\bf X}(:,j)$
belongs to the $k$th cluster,
and ${\bf M}(:,k)$ denotes the centroid of the $k$th cluster.
When $I$ is very large and/or there are redundant features (e.g., highly correlated rows of ${\bf X}$), then it makes sense to perform DR either before or together with clustering. \emph{Reduced $K$-means} (RKM) \cite{de1994k} is a notable joint DR and latent clustering method that is based on the following formulation:
\begin{equation}\label{eq:rkm}
\begin{aligned}
             \min_{{\bf S}\in\mathbb{Z}^{K\times J},{\bf M}\in\mathbb{R}^{F\times K},{\bf P}\in\mathbb{R}^{I\times F}}&\quad\|{\bf X}-{\bf P}{\bf M}{\bf S}\|_F^2\\
						                       {\rm s.t.}&\quad\|{\bf S}(:,j)\|_0=1,~{\bf S}(i,j)\in\{0,1\},\\
																	           &\quad {\bf P}^T{\bf P}={\bf I},
\end{aligned}
\end{equation}
where ${\bf P}$ is a tall semi-orthogonal matrix.
Essentially, RKM aims at factoring ${\bf X}$ into ${\bf H}={\bf M}{\bf S}\in\mathbb{R}^{I\times F}$ and ${\bf P}$,
while enforcing a cluster structure on the columns of ${\bf H}$ -- which is conceptually joint factorization and latent clustering. However, such a formulation loses generality since $F$ (the rank of the model) cannot be smaller than $K$ (the number of clusters);
otherwise, the whole problem is ill-posed.
Note that in practice, the number of clusters and the rank of ${\bf X}$ are not necessarily related; forcing a relationship between them (e.g., $F=K$) can be problematic from an application modeling perspective.
In addition, the cluster structure is imposed as a straight jacket in latent space (no residual deviation, modeled by ${\bf R}$ in ${\bf P}({\bf M}{\bf S}+{\bf R})$ is allowed in \eqref{eq:rkm}), and this is too rigid in some applications.

Another notable formulation that is related to but distinct from RKM is \emph{factorial $K$-means} (FKM) \cite{vichi2001factorial}:
\begin{equation}\label{eq:fkm}
\begin{aligned}
             \min_{{\bf S}\in\mathbb{Z}^{K\times J},{\bf M}\in\mathbb{R}^{F\times K},{\bf P}\in\mathbb{R}^{F\times I}}&\quad\|{\bf P}^T{\bf X}-{\bf M}{\bf S}\|_F^2\\
						                      {\rm s.t.}&\quad\|{\bf S}(:,j)\|_0=1,~{\bf S}(i,j)\in\{0,1\},\\
																	          &\quad {\bf P}^T{\bf P}={\bf I}.
\end{aligned}
\end{equation}
FKM seeks a `good projection' of the data such that the projected data can be better clustered, and essentially performs clustering on ${\bf P}^T{\bf W}{\bf H}$ if ${\bf X}$ admits a low-dimensional factorization as ${\bf X}={\bf W}{\bf H}$. FKM does not force a coupling between $K$ and $F$, and takes the latent modeling error into consideration. On the other hand, FKM ignores the part of ${\bf X}$ that is outside the chosen subspace, so it seeks {\em some} discriminative subspace where the projections cluster well, but ignores variation in the orthogonal complement of ${\bf P}$, since
\begin{align*}
	\|{\bf X}-{\bf P}{\bf M}{\bf S}\|_F^2 &= \|[{\bf P} ~ {\bf P}_\bot]^T({\bf X}-{\bf P}{\bf M}{\bf S})\|_F^2, \\
	& = \|{\bf P}^T{\bf X}-{\bf M}{\bf S}\|_F^2 + \left\|{\bf P}_\bot^T\X \right\|_F^2,
\end{align*}
where ${\bf P}_\bot$ is a basis for the orthogonal complement of ${\bf P}$. So the cost of RKM equals the cost function of FKM plus a penalty for the part of ${\bf X}$ in the orthogonal complement space. FKM was later adapted in \cite{patel2013latent}, where a similar formulation was proposed to combine orthogonal factorization and sparse subspace clustering.

Seeking an orthogonal projection may not be helpful in terms of revealing the cluster structure of ${\bf H}$,
since ${\bf P}^T{\bf W}$ is still a general (oblique) linear transformation that acts on the columns of ${\bf H}$, potentially distorting cluster structure, even if ${\bf P}$ is a basis for ${\bf W}$.

\subsection{Identifiable Factorization Models}\label{ch: iden_cond}
Unlike RMK and FKM that seek an orthogonal factor or projection matrix ${\bf P}$, in this work, we propose to perform latent clustering using \emph{identifiable} low-rank factorization models for matrices and tensors. The main difference in our approach is that we exploit identifiability of the latent factors to help unravel the hidden cluster structure, and in return improve factorization accuracy at the same time. This is sharply different from orthogonal projection models, such as RKM and FKM.
Some important factorization models are succinctly reviewed next.

\subsubsection{Nonnegative Matrix Factorization (NMF)}
Low-rank matrix factorization models are in general non-unique, but nonnegativity can help in this regard \cite{donoho2003does}, \cite{huang2014non}. Loosely speaking, if ${\bf X}={\bf W}{\bf H}$, where ${\bf W}$ and ${\bf H}^T$ are (element-wise) nonnegative and have sufficiently sparse columns and sufficiently spread rows (over the nonnegative orthant), then any nonnegative $(\tilde{\bf W},\tilde{\bf H})$ satisfying ${\bf X}=\tilde{\bf W}\tilde{\bf H}$ can be expressed as
$\tilde{\bf W}={\bf W}{\bm \Pi}{\bf D}$ and $\tilde{\bf H}={\bf D}^{-1}{\bm \Pi}^T {\bf H}$,
where ${\bf D}$ is a full rank diagonal nonnegative matrix and ${\bm \Pi}$ is permutation matrix -- i.e., ${\bf W}$ and ${\bf H}$ are \emph{essentially} unique, or, identifiable up to a common column permutation and scaling-counterscaling.
In practice, NMF is posed as a bilinear optimization problem,
\begin{equation}\label{eq:nmf}
\begin{aligned}
\min_{{\bf W},{\bf H}}&~\|{\bf X}-{\bf W}{\bf H}\|_F^2\\
{\rm s.t.}&~{\bf W}\geq{\bf 0},~{\bf H}\geq{\bf 0}.
\end{aligned}
\end{equation}
NMF is an NP-hard optimization problem. Nevertheless, many algorithms give satisfactory results in practice; see \cite{huang2014putting}. Notice that scaling 

The plain NMF formulation \eqref{eq:nmf} may yield arbitrary nonnegative scaling of the columns of ${\bf W}$ and the rows of ${\bf H}$ due tho the inherent scaling / counter-scaling ambiguity, which can distort distances. This can be avoided by adding a norm-balancing penalty
\begin{equation}\label{eq:nmf_balance}
\begin{aligned}
\min_{{\bf W},{\bf H}}&~\|{\bf X}-{\bf W}{\bf H}\|_F^2 + \mu(\left\| \W\right\|_F^2 +  \left\| \H\right\|_F^2)\\
{\rm s.t.}&~{\bf W}\geq{\bf 0},~{\bf H}\geq{\bf 0}.
\end{aligned}
\end{equation}
It can be easily shown \cite{papalexakis2013k} that an optimum solution of \eqref{eq:nmf_balance} must satisfy $\left\|\W(:,f) \right\|_2 =  \left\|\H(f, :) \right\|_2, ~\forall f$. 

\subsubsection{Volume Minimization (VolMin)-Based Factorization}
NMF has been widely used because it works well in many applications.
If ${\bf W}$ or ${\bf H}$ is dense, however, then the NMF criterion in \eqref{eq:nmf} cannot identify the true factors, which is a serious drawback for several applications. Recent research has shown that this challenge can be overcome using \emph{Volume Minimization} (VolMin)-based structured matrix factorization methods \cite{fu2015blind, fu2016icassp, bioucas2009variable, chan2009convex}
In the VolMin model, the columns of ${\bf H}$ are assumed to live in the unit simplex, i.e., ${\bf 1}^T{\bf H}(:,j)=1$ and ${\bf H}(:,j)\geq{\bf 0}$ for all $j$,
which is meaningful in applications like document clustering \cite{fu2016icassp}, hyperspectral imaging \cite{ma2014signal}, and probability mixture models \cite{hofmann1999probabilistic}.
Under this structural constraint, ${\bf W}$ and ${\bf H}$ are sought via finding
a minimum-volume simplex which encloses all the data columns ${\bf X}(:,j)$.
In optimization form, the VolMin problem can be expressed as follows:
\begin{equation}\label{eq:VolMin}
\begin{aligned}
             \min_{{\bf W}\in\mathbb{R}^{I\times F},{\bf H}\in\mathbb{R}^{F\times J}}&\quad{\rm vol}({\bf W})\\
						                      {\rm s.t.}&\quad{\bf X} ={\bf W}{\bf H}\\
																	          &\quad{\bf H}\geq{\bf 0},~{\bf 1}^T{\bf H}={\bf 1}^T,
\end{aligned}
\end{equation}
where ${\rm vol}(\cdot)$ measures the volume of the simplex spanned by the columns of ${\bf W}$ and is usually a function associated with the determinant of ${\bf W}$ or ${\bf W}^T {\bf W}$ \cite{fu2015blind, fu2016icassp, bioucas2009variable, chan2009convex}.
Notably, it was proven in \cite{fu2015blind} that the optimal solution of \eqref{eq:VolMin} is ${\bf W}{\bm \Pi}$ and ${\bm \Pi}^T{\bf H}$, where ${\bm \Pi}$ is again a permutation matrix, if the ${\bf H}(:,j)$'s are sufficient spread in the probability simplex and ${\bf W}$ is full column rank. Notice that ${\bf W}$ can be dense and even contain negative or complex elements, and still uniqueness can be guaranteed via VolMin.

\subsubsection{Parallel Factor Analysis (PARAFAC)}
For tensor data (i.e., data indexed by more than two indices) low-rank factorization is unique under mild conditions \cite{kruskal1977three, sidiropoulos2000uniqueness}. Low-rank tensor decomposition is known as {\em parallel factor analysis} (PARAFAC) or canonical (polyadic) decomposition (CANDECOMP or CPD). For example, for a three-way tensor $\underline{\bf X}(i,j,l)=\sum_{f=1}^F{\bf A}(i,f){\bf B}(j,f){\bf C}(l,f)$, if
\begin{equation}\label{eq:kruskal}
 k_{\bf A}+k_{\bf B}+k_{\bf C}\geq 2F+2,
\end{equation}
where $k_\A$ is the Kruskal rank of $\A$, if $(\tilde{\bf A},\tilde{\bf B},\tilde{\bf C})$ is such that $\underline{\bf X}(i,j,k)=\sum_{f=1}^F\tilde{\bf A}(i,f)\tilde{\bf B}(j,f)\tilde{\bf C}(k,f)$, then ${\bf A}=\tilde{\bf A}{\bm \Pi}{\bm \Lambda}_1$, ${\bf B}=\tilde{\bf B}{\bm \Pi}{\bm \Lambda}_2$, ${\bf C}=\tilde{\bf C}{\bm \Pi}{\bm \Lambda}_3$,
where ${\bm \Pi}$ is a permutation matrix and $\left\{{\bm \Lambda}_i\right\}_{i=1}^3$ are diagonal scaling matrices such that $\prod_{i=1}^3{\bm \Lambda}_i={\bf I}$. Not that ${\bf A}$, ${\bf B}$ and ${\bf C}$ do not need to be full-column rank for ensuring identifiability.

Making use of the Khatri-Rao product, the tensor factorization problem can be written as
\begin{equation}
\min_{{\bf A},{\bf B},{\bf C}}  ~\left\| {\bf X}_{(1)} -({\bf C}\odot{\bf B}){\bf A}^T\right\|^2_F,
\end{equation}
where ${\bf X}_{(1)}$ is a matrix unfolding of the tensor $\Xt$. There are three matrix unfoldings for this three-way tensor that admit similar model expressions (because one can permute modes and ${\bf A}$, ${\bf B}$, ${\bf C}$ accordingly)
	\begin{align*}
	{\bf X}_{(1)} &= \left[
	\text{vec}(\Xt(1,:,:)), \cdots, \text{vec}(\Xt(I,:,:))\right] \\
	{\bf X}_{(2)} &= \left[
	\text{vec}(\Xt(:,1,:)), \cdots, \text{vec}(\Xt(:,J,:))\right] \\
	{\bf X}_{(3)} &= \left[
	\text{vec}(\Xt(:,:, 1)), \cdots, \text{vec}(\Xt(:,:, K))\right].
	\end{align*}
Like NMF case, PARAFAC is NP-hard in general, but there exist many algorithms offering good performance and flexibility to incorporate constraints, e.g., \cite{xu2013block}, \cite{huang2015flexible}.

Our work brings these factor analysis models together with a variety of clustering tools ranging from $K$-means to $K$-subspace \cite{tseng2000nearest, vidal2010tutorial} clustering to devise novel joint factorization and latent clustering formulations and companion algorithms that outperform the prior art -- including two-step and joint approaches, such as RKM and FKM.

\section{Proposed Approach}
\subsection{Problem Formulation}
Suppose that ${\bf X} \approx{\bf W}{\bf H} \in\mathbb{R}^{I\times J}$, for some element-wise nonnegative ${\bf W}\in\mathbb{R}^{I\times F}$ and ${\bf H}\in\mathbb{R}^{F\times J}$, where the columns of ${\bf H}$ are clustered around $K$ centroids. A natural problem formulation is then
\begin{equation}\label{eq:map}
\begin{aligned}
             \min_{\begin{subarray}{c}{\bf W}\in\mathbb{R}^{I\times F},{\bf H}\in\mathbb{R}^{F\times J}\\{\bf S}\in\mathbb{Z}^{K\times J},{\bf M}\in\mathbb{R}^{F\times K} \end{subarray}}&\quad\|{\bf X}-{\bf W}{\bf H}\|_F^2 + \lambda\|{\bf H}-{\bf M}{\bf S}\|_F^2 \\
						                      {\rm s.t.}&\quad{\bf W}\geq{\bf 0},~{\bf H}\geq{\bf 0},\\
																	          &\quad\|{\bf S}(:,j)\|_0=1,~{\bf S}(k,j)\in\{0,1\},
\end{aligned}
\end{equation}
where the second term is a $K$-means penalty that enforces the clustering prior on the columns of ${\bf H}$, and the tuning parameter $\lambda\geq 0$ balances data fidelity and the clustering prior. This formulation admits a \emph{Maximum A Posteriori} (MAP) interpretation, if ${\bf X}={\bf W}({\bf M}{\bf S}+{\bf E}_2)+{\bf E}_1$, where the data-domain noise ${\bf E}_1$ and the latent-domain noise ${\bf E}_2$ are both drawn from i.i.d. (independent and identically distributed) Gaussian distribution and independent of each other, with variance $\sigma_1^2$ and $\sigma_2^2$, respectively, and $\lambda=\frac{\sigma_1^2}{\sigma_2^2}$.

Assuming that $({\bf W},{\bf H})$ satisfy NMF identifiability conditions, and that ${\bf E}_1$ is negligible (i.e., NMF is exact), ${\bf H}$ will be exactly recovered and thus clustering will be successful as well. In practice of course the factorization model (DR) will be imperfect, and so the clustering penalty will help obtain a more accurate factorization, and in turn better clustering. Note that this approach decouples $K$ and $F$, because it uses a clustering penalty instead of the hard constraint ${\bf H}={\bf M}{\bf S}$ that RKM uses, which results in rank-deficiency when $F < K$.

Formulation \eqref{eq:map} may seem intuitive and well-motivated from a MAP point of view, but there are some caveats. These are discussed next.

\subsection{Design Considerations}
The first problem is scaling. In \eqref{eq:map}, the regularization on ${\bf H}$ implicitly favors a small-norm ${\bf H}$,
since if ${\bf H}$ is small, then simply taking ${\bf M}={\bf 0}$ works. On the other hand, the first term is invariant with respect to scaling of ${\bf H}$, so long as this is compensated in ${\bf W}$. To prevent this, we introduce norm regularization for ${\bf W}$, resulting in
\begin{equation}\label{eq:map_mod}
\begin{aligned}
             \min_{{\bf W},{\bf H},{\bf S},{\bf M}}&\quad\|{\bf X}-{\bf W}{\bf H}\|_F^2 + \lambda\|{\bf H}-{\bf M}{\bf S}\|_F^2 + \eta \|{\bf W}\|_F^2\\
						                      {\rm s.t.}&\quad{\bf W}\geq{\bf 0},~{\bf H}\geq{\bf 0},\\
																	          &\quad\|{\bf S}(:,j)\|_0=1,~{\bf S}(k,j)\in\{0,1\}.
\end{aligned}
\end{equation}
Note that $\|{\bf W}\|_1$ can be used instead of $\|{\bf W}\|_F^2$ to encourage sparsity, if desired.

\begin{figure}
	\centering
	\includegraphics[width=8cm]{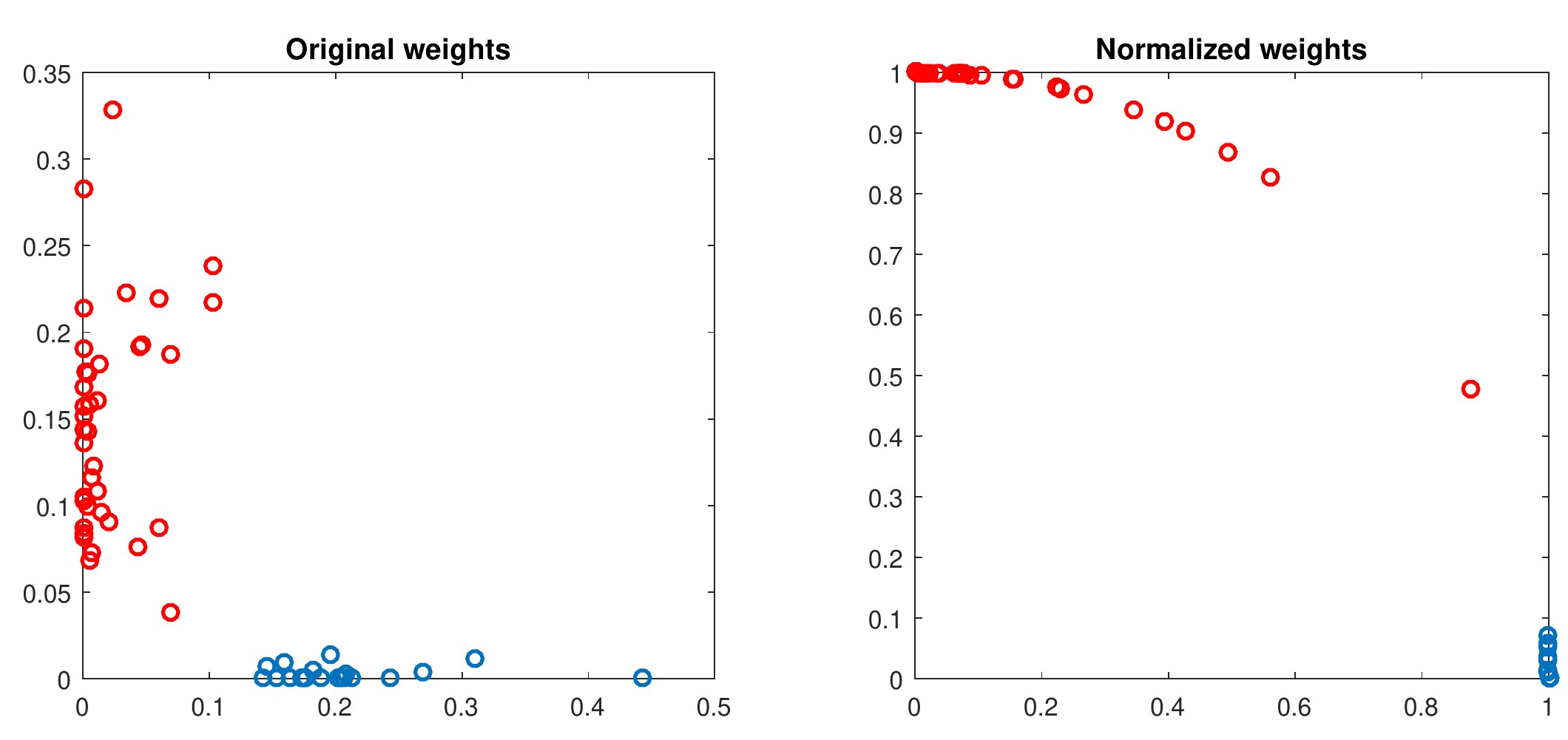}
	\caption{ Normalization in the latent domain helps bringing data points together, creating tight cluster structures. Figure generated by taking a plain NMF on two clusters of documents from Reuters-21578 dataset. Left: 2-dimensional representations of documents; Right: the normalized representations.}
	\label{fig:reu_example}
\end{figure}
Another consideration is more subtle. In many applications, such as document clustering, it has been observed \cite{strehl2000impact,huang2008similarity} that the correlation coefficient or cosine similarity are more appropriate for clustering than Euclidean distance. We have observed that this is also true for latent clustering, which speaks for the need for normalization in the latent domain. Corroborating evidence is provided in Fig.~\ref{fig:reu_example}, which shows the latent representations of two document clusters from the Reuters-25718 dataset. These representations were extracted by plain NMF using $F=2$. In Fig.~\ref{fig:reu_example}, the latent representations on the left are difficult to cluster, especially those close to the origin, but after projection onto the unit $\ell_2$-norm ball  (equivalent to using cosine similarity to cluster the points on the left) the cluster structure becomes more evident on the right.

If $K$-means is applied in the data domain, the cosine similarity metric can be
incorporated easily: by normalizing the data columns using their $\ell_2$-norms in advance,
$K$-means is effectively using cosine dissimilarity as the distance measure.
In our context, however, naive adoption of the cosine similarity for the clustering part can complicate things, since ${\bf H}$ changes in every iteration. To accommodate this, we reformulate the problem as follows.
\begin{equation}\label{nmf-reg-mod}
\begin{aligned}
\min_{\begin{subarray}{c}{\bf W},{\bf H}\\{\bf S},{\bf M},\{d_j\}_{i=1}^J\end{subarray}} & ~\left\| {\bf X} -\mathbf{WH}{\bf D}\right\|^2_F +\lambda\left\| \mathbf{H}-\mathbf{MS}\right\|^2_F+\eta\|{\bf W}\|_F^2\\
{\rm s.t.} &~ \mathbf{W}\geq \mathbf{0},~{\bf H}\geq {\bf 0},~\|{\bf H}(:,j)\|_2 = 1,~\forall j,\\
           &~{\bf D}={\rm Diag}({ d}_1,\ldots,{d}_J),\\
           &~\|{\bf S}(:,j)\|_0=1,~{\bf S}(k,j)\in\{0,1\}.
\end{aligned}
\end{equation}
Introducing the diagonal matrix ${\bf D}$ is crucial: It allows us to fix the columns of $\H$ onto the unit $\ell_2$-norm ball without loss of generality of the factorization model.

The formulation in~\eqref{nmf-reg-mod} can be generalized to tensor factorization models.
Consider a three-way tensor $\underline{\bf X}\in\mathbb{R}^{I\times J \times L}$ with loading factors ${\bf A}\in\mathbb{R}^{I\times F}$, ${\bf B}\in\mathbb{R}^{J\times F}$, ${\bf C}\in\mathbb{R}^{L\times F}$.
Assuming that \emph{rows} of ${\bf A}$ can be clustered into $K$ groups, the \emph{joint tensor factorization and ${\bf A}$-mode latent clustering problem} can be formulated as
\begin{equation}
\begin{aligned}\label{ntf-reg-mod}
\min_{\begin{subarray}{c}{\bf A},{\bf B},{\bf C}\\{\bf S},{\bf M},\{d_\ell\}_{\ell=1}^L\end{subarray}} & ~\left\| {\bf X}_{(1)} -({\bf C}\odot{\bf B})({\bf D}{\bf A})^T\right\|^2_F +\lambda\left\| \mathbf{A}-\mathbf{SM}\right\|^2_F\\
                    &\quad\quad\quad\quad\quad+ \eta\|{\bf B}\|_F^2 + \eta\|{\bf C}\|_F^2\\
{\rm s.t.} &~ \mathbf{A},{\bf B},{\bf C}\geq {\bf 0},~\|{\bf A}(\ell,:)\|_2 = 1,~\forall \ell,\\
           &~{\bf D}={\rm Diag}({d}_1,\ldots,{d}_I),\\
           &~\|{\bf S}(i,:)\|_0=1,~{\bf S}(i,k)\in\{0,1\},~\forall i,k,
\end{aligned}
\end{equation}
where the regularization terms $\|{\bf B}\|_F^2$ and $\|{\bf C}\|_F^2$ are there to control scaling. If one wishes to perform latent clustering in more modes, then norm regularization can be replaced by $K$-means regularization for ${\bf B}$ and/or ${\bf C}$ modes as well. It is still important to have norm regularization for those modes that do not have $K$-means regularization.

An interesting point worth mentioning is that if one adopts VolMin as factorization criterion, then
introducing ${\bf D}$ is not necessary, since VolMin already confines ${\bf H}(:,j)$ on the unit-($\ell_1$-)norm ball.
We also do not need to regularize with the norm of ${\bf W}$, since in this case the scaling of ${\bf W}$ cannot be arbitrary.
This yields
\begin{equation}\label{eq:vol-reg}
\begin{aligned}
\min_{\begin{subarray}{c}{\bf W},{\bf H}\\{\bf S},{\bf M} \end{subarray}} & ~\left\| {\bf X} -\mathbf{WH}\right\|^2_F +\beta\cdot{\rm vol}({\bf W})+\lambda\left\| \mathbf{H}-\mathbf{MS}\right\|^2_F\\
{\rm s.t.} &~ {\bf H}\geq {\bf 0},~ {\bf 1}^T\H = {\bf 1}^T,\\
           &~\|{\bf S}(:,j)\|_0=1,~{\bf S}(k,j)\in\{0,1\}.
\end{aligned}
\end{equation}

\subsection{Extension: Joint Factor Analysis and Subspace Clustering}
In addition to considering Problems~\eqref{nmf-reg-mod}-\eqref{eq:vol-reg},
we also consider their subspace clustering counterparts, i.e., with $K$-means penalties replaced by subspace clustering ones.
Subspace clustering deals with data that come from a union of subspaces \cite{hastie1998metrics}.
Specifically, consider ${\bf X}(:,j)\in{\cal R}({\bf W}(:,{\cal F}_k))$, where ${\cal F}_k$ denotes an index set of a subset of ${\bf W}$'s columns, ${\cal F}_1\cap \cdots \cap{\cal F}_K=\emptyset$ and ${\cal F}_1\cup \cdots \cup{\cal F}_K=\{1,\ldots,K\}$.
Also assume that ${\cal R}({\bf W}(:,{\cal F}_1))\cap\cdots\cap{\cal R}({\bf W}(:,{\cal F}_K))=\{{\bf 0}\}$, which is the independent subspace model \cite{vidal2010tutorial}.
Then, it is evident that
\[{\bf H}(f,{\cal J}_k) ={\bf 0},~f\notin{\cal F}_k,\quad{\bf H}({\cal F}_k,j) ={\bf 0},~ j\notin{\cal J}_k,\]
where ${\cal J}_k$ denotes the set of indices of columns of $\X$ in cluster $k$, i.e. ${\bf X}(:,{\cal J}_k)\in{\cal R}({\bf W}(:,{\cal F}_k))$.

Under this data structure, consider a simple example: if $I=J=F=4$, $K=2$, ${\cal F}_1=\{1,2\}$ and ${\cal F}_2=\{3,4\}$,
then ${\bf H}$ is a block diagonal matrix where the nonzero diagonal blocks are $2\times 2$.
From this illustrative example, it is easy to see that the columns of ${\bf H}$ also come from different subspaces.
The difference is that these subspaces that are spanned by ${\bf H}(:,{\cal J}_k)$ and ${\bf H}(:,{\cal J}_\ell)$
are not only independent, but also orthogonal to each other -- which are much easier to distinguish.
This suggests that performing subspace clustering on ${\bf H}$ is more appealing.

In \cite{patel2013latent}, Patel \etal applied joint dimensionality reduction and subspace clustering on ${\bf P}^T{\bf X}$ using a semi-orthogonal ${\bf P}$, which is basically the same idea as FKM, but using subspace clustering instead of $K$-means as in FKM. However, as we discussed before, when ${\bf W}$ and ${\bf H}$ are identifiable, taking advantage of the identifiability of the factors can further enhance the performance and should therefore be preferred in this case.

Many subspace clustering formulations such as those in \cite{liu2010robust}, \cite{elhamifar2013sparse} can be integrated into our framework, but we limit ourselves to simple $K$-subspace clustering, and assume that the number of subspaces $K$ and subspace dimensions $\{r_i\}_{i=1}^K$ are known, for simplicity. Taking joint NMF and $K$-subspace clustering as an example,
we can express the problem as
\begin{equation}\label{eq:NMF_Ksub}
\begin{aligned}
\min_{\begin{subarray}{c}{\bf W},{\bf H}\\{\bf S},\{{\bm \theta}_{j},{\bm \mu}_k,{\bf U}_k\}\end{subarray}} & ~\left\| {\bf X} -\mathbf{WH}\right\|^2_F+\eta\|{\bf W}\|_F^2 \\ &+ \lambda\sum_{j=1}^J\sum_{k=1}^{K}{\bf S}(k,j)\left\|{\bf H}(:,j)-{\bm \mu}_k-{\bf U}_k{\bm \theta}_{j}\right\|_F^2\\
{\rm s.t.} &~ \mathbf{W}\geq \mathbf{0},~{\bf H}\geq {\bf 0},\\
           &~{\bf U}_k^T{\bf U}_k={\bf I}, ~\forall~ k, \\
           &~\|{\bf S}(:,j)\|_0=1,~{\bf S}(k,j)\in\{0,1\},
\end{aligned}
\end{equation}
where ${\bf S}$ is again a cluster assignment variable,
${\bf U}_k$ denotes an orthogonal basis of the columns of ${\bf H}$ which lie in the $k$th cluster,
${\bm \mu}_k$ is a mean vector of the $k$th cluster,
and ${\bm \theta}_{j} \in \R^{r_k\times 1}$ is the coordinates of the $j$th vector in the subspace.
As in the VolMin case, subspace clustering is also insensitive to the scaling of ${\bf H}(:,j)$,
since the metric for measuring distance to a cluster centroid is the distance to a subspace.
Therefore, the constraints that were added for normalizing ${\bf H}(:,j)$ can be removed.

%

\section{Optimization Algorithms}

In this section, we provide algorithms for dealing with the various problems formulated in the previous section.
The basic idea is alternating optimization -- breaking the variables down to blocks and solving the partial problems one by one.
Updating strategies and convergence properties are also discussed.

\subsection{Joint NMF and $K$-means (JNKM)}
We first consider \eqref{nmf-reg-mod} and \eqref{ntf-reg-mod}.
For ease of exposition, we use \eqref{nmf-reg-mod} as a working example.
Generalization to Problem~\eqref{ntf-reg-mod} is straightforward. Our basic strategy is to alternate between updating ${\bf W}$, ${\bf H}$, ${\bf S}$, ${\bf M}$, and $\{d_i\}_{i=1}^I$ one at a time, while fixing the others. For the subproblems w.r.t. ${\bf S}$ and ${\bf M}$, we propose to use the corresponding (alternating) steps of classical $K$-means \cite{hartigan1979algorithm}.
The minimization w.r.t. $\H$ needs more effort, due to the unit norm and nonegativity constraints.
Here, we propose to employ a variable-splitting strategy. Specifically, we consider the following optimization surrogate:
\begin{equation}\label{nmf-reg-vs}
\begin{aligned}
&\min_{{\bf W},{\bf H},{\bf Z},{\bf S},{\bf M},\{d_i\}_{i=1}^I}~ \left\| {\bf X} -\mathbf{WH}{\bf D}\right\|^2_F +\lambda\left\| \mathbf{H}-\mathbf{MS}\right\|^2_F\\ &\quad\quad\quad\quad  \quad   \quad\quad\quad\quad\quad\quad+\eta\|{\bf W}\|_F^2 + \mu\|{\bf H}-{\bf Z}\|_F^2\\
&\quad\quad\quad{\rm s.t.}\quad     {\bf W}\geq \mathbf{0},~{\bf H}\geq {\bf 0},~\|{\bf Z}(:,j)\|_2 = 1,~\forall j,\\
&\quad\quad\quad \quad  \quad ~     {\bf D}={\rm Diag}({ d}_1,\ldots,{d}_J),\\
&\quad\quad\quad \quad \quad  ~      \|{\bf S}(:,i)\|_0=1,~{\bf S}(k,j)\in\{0,1\},~\forall k,j,
\end{aligned}
\end{equation}
where $\mu\geq 0$ and ${\bf Z}$ is a slack variable. Note that ${\bf Z}$ is introduced to `split' the effort of dealing with $\mathbf{H}\geq \mathbf{0}$ and $\|{\bf H}(:,j)\|_2 = 1$
in two different subproblems.
Notice that when $\mu=+\infty$, \eqref{nmf-reg-vs} is equivalent to \eqref{nmf-reg-mod};
in practice, a large $\mu$ can be employed to enforce ${\bf H}\approx {\bf Z}$.

Problem~\eqref{nmf-reg-vs} can be handled as follows.
First, ${\bf H}$ can be updated by solving
\begin{align}
{\bf H}\leftarrow\arg\min_{{\bf H}\geq{\bf 0}}~&\left\| {\bf X} -\mathbf{WH}{\bf D}\right\|^2_F +\lambda\left\| \mathbf{H}-\mathbf{MS}\right\|^2_F \nonumber\\
                                       &+ \mu\|{\bf H}-{\bf Z}\|_F^2, \label{eq:H_update}
\end{align}
which can be easily converted to a nonnegative least squares (NLS) problem, and efficiently solved to optimality by many existing methods.
Here, we employ an alternating direction method of multipliers (ADMM)-based \cite{boyd2011distributed}
algorithm to solve Problem~\eqref{eq:H_update}.
The update of ${\bf W}$, i.e.,
\begin{align}\label{eq:w_update}
{\bf W}\leftarrow\arg\min_{{\bf W}\geq{\bf 0}}~\left\| {\bf X} -\mathbf{WH}{\bf D}\right\|^2_F+\eta\|{\bf W}\|_F^2,
\end{align}
is also an NLS problem.
The subproblem w.r.t. $d_j$ admits closed-form solution,
\begin{align}\label{eq:d_update}
d_j \leftarrow {\bf b}_j^T {\bf X}(:,j)/({\bf b}_j^T{\bf b}_j),
\end{align}
where ${\bf b}_j={\bf W}{\bf H}(:,j)$. The update of ${\bf Z}(:,j)$ is also closed-form,
\begin{align}\label{eq:Z_update}
{\bf Z}(:,j)\leftarrow\frac{{\bf H}(:,j)}{\|{\bf H}(:,j)\|_2}.
\end{align}
The update of ${\bf M}$ comes from the $K$-means algorithm. Let ${\cal J}_k=\{j~|~{\bf S}(k,j)=1\}$. Then
\begin{equation}\label{eq:M_update}
{\bf M}(:,k) \leftarrow\frac{\sum_{j\in{\cal J}_k}{\bf H}(:,j)}{|{\cal J}_k|}.
\end{equation}
The update of ${\bf S}$ also comes from the $K$-means algorithm
\begin{equation}\label{eq:S_update}
{\bf S}(\ell,j) \leftarrow \begin{cases} 1,&\quad \ell =\arg\min_{k}~\|{\bf H}(:,j)-{\bf M}(:,k)\|_2\\ 0,&\quad {\rm otherwise}. \end{cases}
\end{equation}
The overall algorithm alternates between updates \eqref{eq:H_update}-\eqref{eq:S_update}.

\subsection{Joint NTF and $K$-means (JTKM)}
As in the JNKM case, we employ variable splitting and introduce a $\Z$ variable to \eqref{ntf-reg-mod}
\begin{equation}
\begin{aligned}\label{ntf-reg-var}
\min_{\begin{subarray}{c}{\bf A},{\bf B},{\bf C}\\{\bf S},{\bf M},\{d_\ell\}_{\ell=1}^L\end{subarray}} & ~\left\| {\bf X}_{(1)} -({\bf C}\odot{\bf B})({\bf D}{\bf A})^T\right\|^2_F +\lambda\left\| \mathbf{A}-\mathbf{SM}\right\|^2_F\\
&~+ \eta(\|{\bf B}\|_F^2 + \|{\bf C}\|_F^2) + \mu\left\|\A -\Z \right\|_F^2 \\
{\rm s.t.} &~ \mathbf{A},{\bf B},{\bf C}\geq {\bf 0},~\|{\bf Z}(\ell,:)\|_2 = 1,~\forall \ell,\\
&~{\bf D}={\rm Diag}({d}_1,\ldots,{d}_I),\\
&~\|{\bf S}(i,:)\|_0=1,~{\bf S}(i,k)\in\{0,1\},~\forall i,k,
\end{aligned}
\end{equation}
The algorithm for dealing with Problem~\eqref{ntf-reg-var} is similar to that of the NMF case.
By treating ${\bf X}_{(1)}$ as ${\bf X}$,
$({\bf B}\odot{\bf C})$ as ${\bf W}$ and ${\bf A}^T$ as ${\bf H}$, the updates of ${\bf A}$, ${\bf D}$, $\Z$, ${\bf S}$ and ${\bf M}$ are the same as those in the previous section.
To update ${\bf B}$ and ${\bf C}$, we make use of the other two matrix unfoldings
\begin{subequations}
\begin{align}
       {\bf B}&\leftarrow \arg\min_{\B\geq {\bf 0}} \left\| {\bf X}_{(2)}-(\C\odot \D\A)\B^T\right\| _F^2 + \eta\left\|\B \right\|_F^2, \\
			 {\bf C}&\leftarrow \arg\min_{\C\geq {\bf 0}} \left\|  {\bf X}_{(3)}-(\B\odot \D\A)\C^T\right\| _F^2 + \eta\left\|\C \right\|_F^2.
\end{align}
\end{subequations}
These are nonnegative linear least squares problems, and thus can be efficiently solved.

\subsection{Joint VolMin and $K$-means (JVKM)}

For VolMin-based factorization, one major difficulty is dealing with the volume measure ${\rm vol}({\bf W})$, which is usually defined as ${\rm vol}({\bf W}) = \det({\bf W}^T{\bf W})$ \cite{bioucas2009variable, fu2015blind,fu2016icassp}.
If clustering is the ultimate objective, however, we can employ more crude volume measures for the sake of computational simplicity. With this in mind, we propose to employ the following approximation of simplex volume \cite{berman2004ice}:
${\rm vol}({\bf W})\approx\sum_{f=1}^F{\sum_{\ell>f}^F}\|{\bf W}(:,f)-{\bf W}(:,\ell)\|_2^2={\rm Tr}({\bf W}{\bf G}{\bf W}^T)$, where ${\bf G}= F{\bf I} - {\bf 1}{\bf 1}^T$.
The regularizer ${\rm Tr}({\bf W}{\bf G}{\bf W}^T)$ measures the volume of the simplex that is spanned by the columns of ${\bf W}$
by simply measuring the distances between the vertices.
This approximation is coarse, but reasonable.
Hence, Problem~\eqref{eq:vol-reg} can be tackled using a four-block BCD, i.e.,
\begin{subequations}
\begin{align}
           {\bf W}& \leftarrow\arg\min_{\bf W}~~\left\| {\bf X} -\mathbf{WH}\right\|^2_F +\beta{\rm Tr}({\bf W}{\bf G}{\bf W}^T), \label{eq:VolMin_M}\\
					 {\bf H}& \leftarrow\arg\min_{{\bf 1}^T{\bf H}={\bf 1}^T,{\bf H}\geq{\bf 0}}~~~\left\| {\bf X} -\mathbf{WH}\right\|^2_F + \lambda\|{\bf H}-{\bf M}{\bf S}\|_F^2 \label{eq:VolMin_H}					
\end{align}
\end{subequations}
Note that Problem~\eqref{eq:VolMin_M} is a convex quadratic problem that has closed-form solution,
and Problem~\eqref{eq:VolMin_H} is a simplex-constrained least squares problem that can be solved efficiently via many solvers. The updates for $\M$ and $\S$ are still given by \eqref{eq:M_update} and \eqref{eq:S_update}.

\subsection{Joint NMF and $K$-subspace (JNKS)}


Similar to JNKM, the updates of $\W, \H$ in \eqref{eq:NMF_Ksub} are both NLS problems, and can be easily handled.
To update the subspace and the coefficients, we need to solve a $K$-subspace clustering problem \cite{vidal2010tutorial}
\begin{equation}\label{eq:Ksub_U}
\begin{aligned}
\min_{\{\U_k\}, \{{\bm \mu}_k\}, \{\boldsymbol{\theta}_j\}} & ~\sum_{j=1}^J\sum_{k=1}^{K}{\bf S}(k,j)\left\|{\bf H}(:,j)-{\bm \mu}_k-{\bf U}_k{\bm \theta}_{j}\right\|_F^2\\
{\rm s.t.} &~ {\bf U}_k^T{\bf U}_k={\bf I}, ~\forall~ k.
\end{aligned}
\end{equation}
Let $\H(:, {\cal F}_k)$ denote the data points in cluster $k$, and $\boldsymbol{\Theta}_k := \{\boldsymbol{\theta}_j | j\in {\cal F}_k\}$. We can equivalently write \eqref{eq:Ksub_U} as
\begin{equation}\label{eq:Ksub_U_mod}
\begin{aligned}
\min_{\{\U_k\}, \{{\bm \mu}_k\}, \{\boldsymbol{\Theta}_k\}} & ~\sum_{k=1}^{K}\left\|\H(:, {\cal F}_k)-{\bm \mu}_k{\bf 1}^T-{\bf U}_k\boldsymbol{\Theta}_k\right\|_F^2\\
{\rm s.t.} &~ {\bf U}_k^T{\bf U}_k={\bf I}, ~\forall~ k.\end{aligned}
\end{equation}
It can be readily seen that the update of each subspace is independent of the others. For cluster $k$, we first remove its center, and then take a SVD,
\begin{subequations}
	\begin{align}\label{svd}
	{\bm \mu}_k &\leftarrow \frac{\H(:, {\cal F}_k) {\bf 1}}{\left\| \S(k, :)\right\|_0 } ,\\
	[\widehat{\U}, \widehat{\boldsymbol{\Sigma}}, \widehat{\V}^T ] &\leftarrow \text{svd} (\H(:, {\cal F}_k) - \boldsymbol{\mu}_k{\bf 1}^T),\\
	\U_k &\leftarrow \widehat{\U}(:,1:r_k),\\
	\boldsymbol{\Theta}_k & \leftarrow \widehat{\boldsymbol{\Sigma}}(1:r_k,1:r_k) \widehat{\V}(:,1:r_k)^T.
	\end{align}
\end{subequations}
To update the subspace assignment $\S$, we solve
\begin{equation}\label{eq:Ksub_S}
\begin{aligned}
\min_{{\bf S}} & ~\sum_{j=1}^J\sum_{k=1}^{K}{\bf S}(k,j)\left\|{\bf H}(:,j)-{\bm \mu}_k-{\bf U}_k{\bm \theta}_{j}\right\|_F^2\\
{\rm s.t.} &~ \|{\bf S}(:,j)\|_0=1,~{\bf S}(k,j)\in\{0,1\}.
\end{aligned}
\end{equation}
With $\text{dist}(j,k) := \left\| (\I - \U_k\U_k^T)(\H(:, j) - \boldsymbol{\mu}_k)\right\|_2$, the update of $\S$ is given by \cite{vidal2010tutorial}
\begin{align}
\S(\ell,j) \leftarrow \begin{cases}
1 &\quad \ell = \text{argmin}_{k}~\text{dist}(j,k)\\
0 &\quad \text{otherwise}.
\end{cases}
\end{align}

\subsection{Convergence Properties}
The proposed algorithms, i.e., JNKM, JVKM, JTKM, and their subspace clustering counterparts,
share some similar traits. Specifically, all algorithms solve the conditional updates (block minimization problems)
optimally. From this it follows that
\begin{Property}
JNKM, JTKM, and JVKM, yield a non-increasing cost sequence in terms of their respective cost functions in \eqref{nmf-reg-vs}, \eqref{ntf-reg-var}, and \eqref{eq:vol-reg}, respectively. The same property is true for their subspace clustering counterparts.
\end{Property}
Monotonicity is important in practice -- it ensures that an algorithm makes progress in every iteration towards the corresponding design objective. In addition, it leads to convergence of the cost sequence when the cost function is bounded from below (as in our case), and such convergence can be used for setting up stopping criteria in practice.

In terms of convergence of the iterates (the sequence of `interim' solutions), when using a cyclical block updating strategy all algorithms fall under the Gauss-Seidel block coordinate descent (BCD) framework, however related convergence results \cite{bertsekas1999nonlinear} cannot be applied because some blocks involve {\em nonconvex} constraints. If one uses a more expensive (non-cyclical) update schedule, then the following holds.
\begin{Property}
If the blocks are updated following the maximum block improvement (MBI) strategy (also known as the Gauss-Southwell rule), then every limit point of the solution sequence produced by JNKM, JTKM, and JVKM is a block-wise minimum of \eqref{nmf-reg-vs}, \eqref{ntf-reg-var}, and \eqref{eq:vol-reg}, respectively. A block-wise minimum is a point where it is not possible to reduce the cost by changing any single block while keeping the others fixed.
\end{Property}
The MBI update rule \cite{chen2012maximum} is similar to BCD, but it does not update the blocks cyclically. Instead, it tries updating each block in each iteration, but actually updates only the block that brings the biggest cost reduction. The MBI result can be applied here since we solve each block subproblem to optimality. The drawback is that MBI is a lot more expensive per iteration. If one is interested in obtaining a converging subsequence, then a practical approach is to start with cyclical updates, and then only use MBI towards the end for `last mile' refinement. We use Gauss-Seidel in our simulations, because it is much faster than MBI.

\section{Synthetic Data Study}

In this section, we use synthetic data to explore scenarios where the proposed algorithms show promising performance relative to the prior art. We will consider real datasets that are commonly used as benchmarks in the next section.

Our algorithms simultaneously {\em estimate} the latent factors and {\em cluster}, so we need a metric to assess estimation accuracy, and another for clustering accuracy. For the latter, we use the ratio of correctly clustered points over the total number of points (in $[0,1]$, the higher the better) \cite{cai2011graph,cai2011locally}. Taking into account the inherent column permutation and scaling indeterminacy in estimating the latent factor matrices, estimation accuracy is measured via the following {\em matched} mean-square-error measure (MSE, for short)
\begin{equation*}
{\rm MSE}=
\min_{ \substack{ \bm{\pi} \in \Pi, \\ c_1,\ldots,c_F \in \{ \pm 1
		\}  } }
\frac{1}{F} \sum_{f=1}^F \left\| \frac{ {\bf W}(:,f) }{ \| {\bf W}(:,f) \|_2 } - c_f   \frac{ \hat{\bf W}(:,{\pi_f}) }{ \| \hat{\bf W}(:,{\pi_f}) \|_2 }    \right\|_2^2, 
\label{eq:MSEdef}
\end{equation*}
where ${\bf W}$ and $\hat{\bf W}$ represents the ground truth factor and the estimated factor, respectively, $\Pi$ is the set of all permutations of the set $\left\{1,\cdots,F\right\}$, $\bm{\pi}=[\pi_1,\cdots,\pi_F]^T$ is there to resolve the permutation ambiguity, and $c_f$ to resolve the sign ambiguity when there is no nonnegativity constraint.

\subsection{Joint Matrix Factorization and Latent $K$-means}
We generate synthetic data according to
\begin{align}\label{eq: data_gen}
\X &= \W(\M\S + {\bf E_2}) + {\bf E_1},\\
&= \W\H + {\bf E_1},\nonumber
\end{align}
where $\X$ is the data matrix,
$\W$ is the basis matrix,
$\H=\M\S + {\bf E_2}$ is the factor where the clustering structure lies in,
$\M$ and $\S$ are the centroid matrix and the assignment matrix, respectively,
and ${\bf E}_i$ for $i=2,1$ denote the modeling errors and the measurement noise, respectively.
We define the Signal to Noise Ratio (SNR) in the data and latent space, respectively, as $\text{SNR}_1 = \frac{\left\| \W\H\right\|_F^2 }{\left\| {\bf E}_1\right\|_F^2 }$ and $\text{SNR}_2 = \frac{\left\|\M\S \right\|_F^2 }{\left\| {\bf E}_2\right\|_F^2}$. SNR$_1$ is the `conventional' SNR that characterizes the data corruption level,
and SNR$_2$ is for characterizing the latent domain modeling errors. All the simulation results are obtained by averaging 100 Monte Carlo trials.

We employ several clustering and factorization algorithms as baselines, namely, the original $K$-means, the reduced $K$-means (RKM), the factorial $K$-means algorithm (FKM), and a simple two-stage combination of nonnegative matrix factorization (NMF) and $K$-means (NMF-KM).

We first test the algorithms under an NMF model.
Specifically, ${\bf W}\in\mathbb{R}^{I\times F}$ is drawn from i.i.d. standard Gaussian distribution, with all negative entries set to zero. ${\bf H}$ is generated with the following steps:
\begin{enumerate}
	\item Generate ${\M}\in\mathbb{R}^{F\times K}$ by setting the first $F$ columns as unit vectors (to ensure identifiability of the NMF model), i.e. $\M(:, 1:F) = \I$, and entries in the remaining $K-F$ columns are randomly generated from an i.i.d. uniform distribution in $[0, 1]$; set $\S \in \mathbb{R}^{K\times J}$ as $\S = [\I, \I, \cdots, \I]$;
	
	\item Draw ${\bf E}_2$ from an i.i.d. standard Gaussian distribution, set $\H \leftarrow \M\S + {\bf E}_2$ and perform the following steps
	\begin{subequations}
		\begin{align}
		\H &\leftarrow (\H)_+, \label{eq: generate_H_1}\\
		{\bf E}_2 &\leftarrow \H - \M\S, \\
		{\bf E}_2 &\leftarrow \gamma {\bf E}_2 \label{eq: generate_H_3}, \\
		\H &\leftarrow \M\S + {\bf E}_2 \label{eq: generate_H_4},
		\end{align}
	\end{subequations}	
	where $\gamma$  in \eqref{eq: generate_H_3} is a scaling constant determined by the desired $\text{SNR}_2$, and $(\cdot)_+$ takes the nonnegative part of its argument. We may need to repeat the above steps \eqref{eq: generate_H_1}$\sim$\eqref{eq: generate_H_4} several times, till we get a nonnegative $\H$ with desired $\text{SNR}_2$ (in our experience, usually one repetition suffices).
\end{enumerate}
We then multiply $\W$ and $\H$ and add ${\bf E}_1$, where ${\bf E}_1$ is drawn from an i.i.d. standard Gaussian distribution and scaled for the desired $\text{SNR}_1$. With this process, ${\bf W}$ and ${\bf H}$ are sparse and identifiable (when ${\bf E}_1 = {\bf 0}$) with very high probability \cite{donoho2003does,huang2014non}.
Finally, we replace 3\% of the columns of $\X$ with all-ones vectors that act as outliers, which are common in practice.


Table~\ref{simu: large} presents the clustering accuracies of various algorithms using $I=50$, $J=1000$, $F=7$, and $K=10$.
The MSEs of the estimated $\hat{\bf W}$ of the factorization-based algorithms are also presented.
We set the parameters of the JNKM method to be $\lambda=1$, $\mu=100$ and $\eta=10^{-1}$.
Here, SNR$_1$ is fixed to be 15 dB and SNR$_2$ varies from 3 dB to 18 dB. The JNKM is initialized with an NMF solution \cite{kim2011fast}, and the JVKM is initialized with the SISAL \cite{bioucas2009variable} algorithm.
We see that, for all the SNR$_2$'s under test, the proposed JNKM yields the best clustering accuracies.
RKM and FKM give poor clustering results since they cannot resolve the distortion brought by ${\bf W}$, as we discussed before. Our proposed method works better than NMF-KM, since JNKM estimates ${\bf H}$ more accurately (cf. the MSEs of the estimated factor ${\bf W}$) -- by making use of the cluster structure on ${\bf H}$ as prior information.

Note that in order to make the data fit the VolMin model, we normalized the columns of $\X$ to unit $\ell_1$-norm as suggested in \cite{gillis2014robust}. Due to noise, such normalization cannot ensure that the data follow the VolMin model exactly, however we observe that the proposed JVKM formulation still performs well in this case.

\begin{table}
	\scriptsize
	\centering	
	\caption{Clustering and factorization accuracy for identifiable NMF vs. $\text{SNR}_2$, for $I = 50, J = 1000, F = 7, K = 10$, $\text{SNR}_1 = 15$ dB.}
	\resizebox{0.49\textwidth}{!}{%
		\begin{tabular}{c|c||c|c|c|c|c|c}
			\cline{1-8}
			\hline
			\hline
			\multicolumn{2}{c||}{$\text{SNR}_2$ [dB]}
			& 3     & 6     & 9     & 12    & 15    & 18  \\
			\hline
			\hline
			\multicolumn{1}{ c|  }{\multirow{6}{*}{AC[\%]} } & {KM} & 77.43 & 81.5  & 82.9  & 81.47 & 82.68 & 84.5     \\
			\cline{2-8}				
			\multicolumn{1}{ c|  }{} & {RKM} & 77.51 & 76.62 & 73.71 & 72.43 & 71.35 & 71.63     \\
			\cline{2-8}	
			\multicolumn{1}{ c|  }{} & {FKM} & 15.12 & 15.68 & 16.6  & 17.14 & 37.5  & 59.74        \\
			\cline{2-8}		
			\multicolumn{1}{ c|  }{} & {JNKM } & 	\textbf{88.1} & \textbf{95.12} & \textbf{96.51} & \textbf{96.13} & \textbf{96.43} & \textbf{95.65}     \\
			\cline{2-8}		
			\multicolumn{1}{ c|  }{} & {JVKM } & 	75.84 & 83.87 & 87.87 & 89.96 & 90.27 & 89.36     \\
			\cline{2-8}
			\multicolumn{1}{ c|  }{} & {NMF-KM } & 84.72 & 86.62 & 88.96 & 90.95 & 90.87 & 92.34        \\ 		
			\cline{1-8}	
			\multicolumn{1}{ c|  }{\multirow{3}{*}{MSE[dB]} } & {JNKM }&
			\textbf{-28.09} & \textbf{-27.82} & \textbf{-27.54} & \textbf{-26.59} & \textbf{-26.91} & \textbf{-26.26}     \\
			\cline{2-8}		
			\multicolumn{1}{ c| }{} & {JVKM } & -16.41 & -16.98 & -16.37 & -15.61 & -15.19 & -14.9        \\
			\cline{2-8}		
			\multicolumn{1}{ c| }{} & {NMF-KM } & -27.09 & -26.75 & -26.7 & -25.58 & -26.05 & -25.31      \\
			\hline
			\multicolumn{1}{ c|  }{\multirow{6}{*}{TIME[s]} } &  {KM}  & \textbf{0.14} & \textbf{0.05} & \textbf{0.05} & \textbf{0.07} & \textbf{0.06} & \textbf{0.06}      \\
			\cline{2-8}		
			\multicolumn{1}{ c|  }{} & {RKM} &
			0.18  & 0.13  & 0.16  & 0.17  & 0.17  & 0.18 \\
			\cline{2-8}	
			\multicolumn{1}{ c|  }{} & {FKM} & 1.45  & 0.59  & 0.64  & 0.79  & 0.82  & 0.56 \\
			\cline{2-8}	
			\multicolumn{1}{ c|  }{} & {JNKM} & 3.37  & 2.99  & 3.13  & 3.22  & 3.06  & 3.36     \\
			\cline{2-8}	
			\multicolumn{1}{ c|  }{} & {JVKM} & 5.7   & 5.06  & 4.73  & 4.53  & 4.42  & 4.45     \\
			\cline{2-8}	
			\multicolumn{1}{ c|  }{} & {NMF-KM } & 0.76  & 0.68  & 0.73  & 0.83  & 0.75  & 0.86  \\	
			\cline{1-8}				
		\end{tabular}	
		\label{simu: large}
	}
\end{table}

\begin{table}
	\scriptsize
	\centering	
	\caption{Clustering and factorization accuracy for identifiable NMF vs. $\text{SNR}_1$, for $I = 50, J = 1000, F = 7, K = 10$, $\text{SNR}_2 = 10$ dB.}
	\resizebox{0.49\textwidth}{!}{%
		\begin{tabular}{c|c||c|c|c|c|c|c}
			\cline{1-8}
			\hline
			\hline
			\multicolumn{2}{c||}{$\text{SNR}_1$ [dB]}
			& 5     & 10    & 15    & 20    & 25    & 30   \\
			\hline	
			\hline
			\multicolumn{1}{ c|  }{\multirow{6}{*}{AC[\%]} } & {KM} & 78.65 & 77.89 & 82.89 & 84.53 & 88.43 & 86.97      \\
			\cline{2-8}				
			\multicolumn{1}{ c|  }{} & {RKM} & 79.54 & 72.84 & 72.87 & 71.15 & 71.37 & 72.06      \\
			\cline{2-8}	
			\multicolumn{1}{ c|  }{} & {FKM} & 17.91 & 17.3  & 16.76 & 16.68 & 16.44 & 16.51         \\
			\cline{2-8}		
			\multicolumn{1}{ c|  }{} & {JNKM } & 	\textbf{93.28} & \textbf{94.69} & \textbf{95.73} & \textbf{96.33} & \textbf{96.43} & \textbf{96.04}      \\
			\cline{2-8}		
			\multicolumn{1}{ c|  }{} & {JVKM } & 	71.78 & 82.74 & 87.43 & 91.68 & 92.43 & 93.17     \\
			\cline{2-8}
			\multicolumn{1}{ c|  }{} & {NMF-KM } & 84.95 & 86.74 & 89.57 & 90.87 & 90.66 & 91.61      \\ 		
			\cline{1-8}	
			\multicolumn{1}{ c|  }{\multirow{3}{*}{MSE[dB]} } & {JNKM }&
			\textbf{-18.03} & \textbf{-23.95} & \textbf{-26.71} & -27.17 & -27.09 & -26.19      \\
			\cline{2-8}		
			\multicolumn{1}{ c| }{} & {JVKM } & -4.66 & -12.19 & -15.96 & -20.21 & -25.14 & \textbf{-31.05}         \\
			\cline{2-8}		
			\multicolumn{1}{ c| }{} & {NMF-KM } & -17.63 & -23.36 & -26.29 & \textbf{-27.48} & \textbf{-27.76} & -27.13       \\
			\hline
			\multicolumn{1}{ c|  }{\multirow{6}{*}{TIME[s]} } &  {KM}  & \textbf{0.08} & \textbf{0.08} & \textbf{0.05} & \textbf{0.05} & \textbf{0.04} & \textbf{0.04}      \\
			\cline{2-8}		
			\multicolumn{1}{ c|  }{} & {RKM} &
			0.18  & 0.18  & 0.16  & 0.14  & 0.11  & 0.11 \\
			\cline{2-8}	
			\multicolumn{1}{ c|  }{} & {FKM} & 0.68  & 0.7   & 0.66  & 0.69  & 0.61  & 0.67  \\
			\cline{2-8}	
			\multicolumn{1}{ c|  }{} & {JNKM} & 3.1   & 3.29  & 3     & 3.13  & 2.99  & 2.92    \\
			\cline{2-8}	
			\multicolumn{1}{ c|  }{} & {JVKM} & 3.21  & 3.83  & 4.57  & 5.12  & 5.01  & 4.94     \\
			\cline{2-8}	
			\multicolumn{1}{ c|  }{} & {NMF-KM } & 0.74  & 0.75  & 0.7   & 0.66  & 0.64  & 0.62   \\	
			\cline{1-8}			
		\end{tabular}	
		\label{simu: large_snr1}
	}
\end{table}

\begin{figure}
	\centering
	\includegraphics[width=0.5\textwidth]{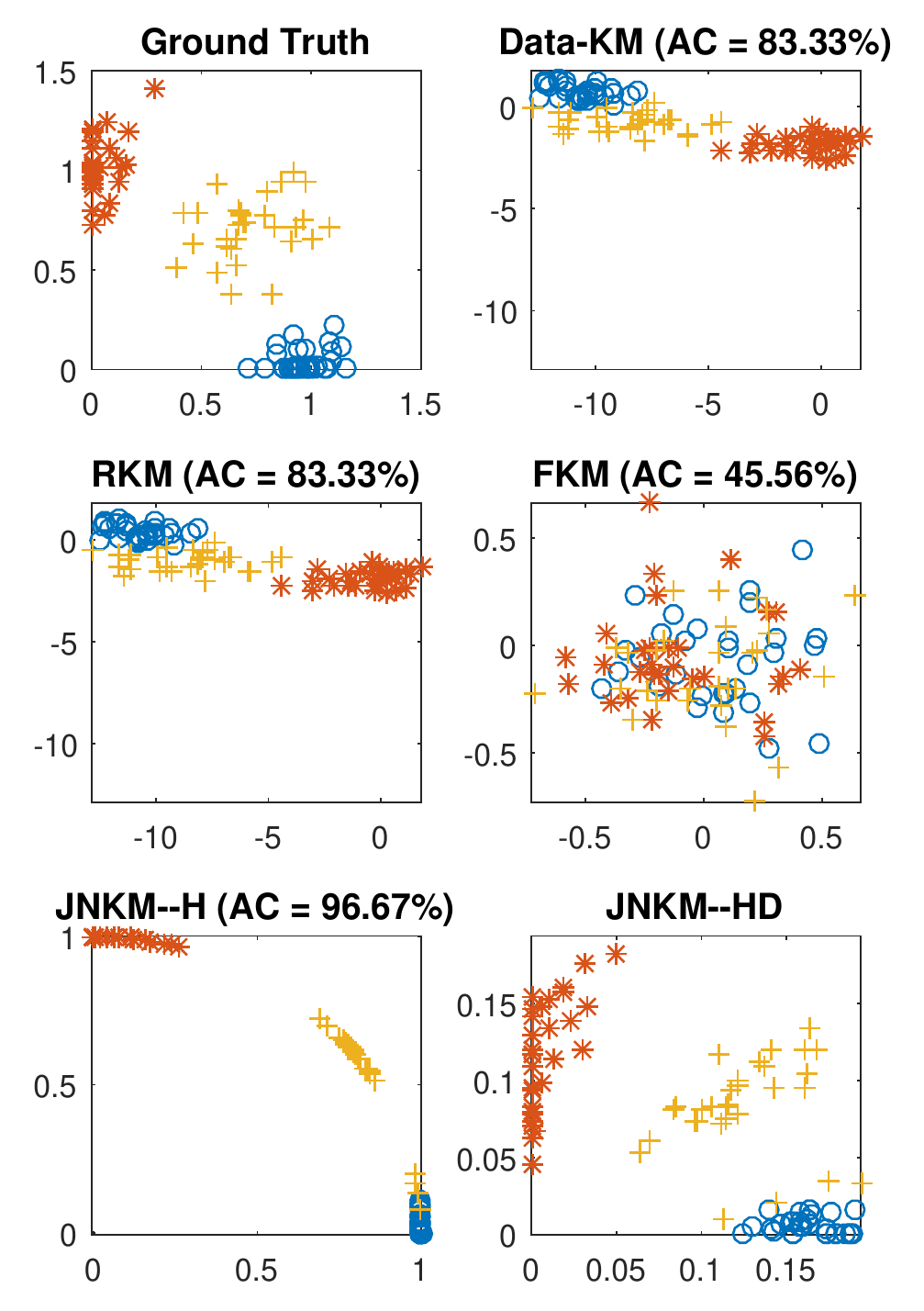}
	\caption{Illustration of how linear transformation obscures the latent cluster structure, and how identifiable models can recover this cluster structure. Top left: true latent factor $\H$; Top right: data domain $\X = \W\H +{\bf E}_1$, visualized using SVD (two principal components); Middle left: projected data found by RKM, ${\bf P}^T\X$; Middle right: projected data found by FKM, ${\bf P}^T\X$; Bottom left: $\H$ found by JNKM; Bottom right: $\H\D$ found by JNKM. In the top right subfigure, the clustering accuracy of running K-means directly on the data is shown; for other figures, the clustering accuracy given by corresponding method is shown.}
	\label{fig:visual_example}
\end{figure}

To better understand the reason why our method performs well, we present an illustrative example where the latent factor $\H$ lies in a two-dimensional subspace (so that it can be visualized), and has a clear cluster structure. The basis $\W$ is an $8\times 2$ matrix. The factor are generated such that the NMF model is identifiable.
Fig.~\ref{fig:visual_example} shows the true latent factors, together with those found by various methods.
Clearly, ${\bf W}$ brings some distance distortion to the cluster structure in ${\bf H}$ (cf. top right subfigure).
We see from this example that if the factorization model is identifiable, using the proposed approach helps greatly in removing the distance distortion brought by ${\bf W}$, as indicated by the last row of Fig.~\ref{fig:visual_example}.
On the other hand, the other semi-orthogonal projection-based algorithms do not have this salient feature.

{Table~\ref{simu: large_snr1} presents the results under various SNR$_1$'s.
Here, we fix SNR$_2=10$ dB, and the other settings are the same as in the previous simulation.
We see that the clustering accuracies are not so sensitive to SNR$_1$, and the proposed JNKM outperforms other methods in AC and MSE in most of the cases.} Table~\ref{simu: clus} presents the ACs and MSEs for fixed rank $F=7$ as the number of clusters $K$ varies from $5$ to $11$. Here $I = 50$, $J=100K$, $\text{SNR}_1 = 6 \text{dB}$, and $\text{SNR}_2 = 8\text{dB}$. We observe
that the performance of all methods degrades when we add more clusters, which is expected. However, RKM and FKM suffer more than the proposed method.

\begin{table}
	\scriptsize
	\centering	
	\caption{Simulation comparison of the clustering methods, identifiable NMF model. $I = 50$, $F = 7$, $J = 100K$.}
	\resizebox{.49\textwidth}{!}{
		\begin{tabular}{c|c||c|c|c|c|c|c|c}
			\cline{1-8}
			\hline
			\hline
			\multicolumn{2}{c||}{$K$}
			& 5     & 6     & 7     & 8     & 9     & 10    & 11   \\
			\hline	
			\hline			
			\multicolumn{1}{ c|  }{\multirow{3}{*}{AC[\%]} } & {RKM} & 79.97 & 78.57 & 76.6  & 76.16 & 75.48 & 75.22 & 75.54    \\
			\cline{2-9}	
			\multicolumn{1}{ c|  }{} & {FKM} & 47.75 & 34.01 & 27.32 & 21.96 & 18.45 & 16.55 & 15.14      \\
			\cline{2-9}		
			\multicolumn{1}{ c|  }{} & {JNKM } &
			\textbf{97.6} & \textbf{97.5} & \textbf{97.43} & \textbf{97.37} & \textbf{96.88} & \textbf{96.63} & \textbf{95.48}  \\ 	
			\hline
			\multicolumn{1}{ c|  }{ MSE[dB] } & {JNKM }&
			-4.7  & -7.52 & -25.08 & -24.77 & -24.3 & -24.17 & -23.81     \\
			\hline	
		\end{tabular}	
		\label{simu: clus}
	}
\end{table}

\begin{figure}[h!]
	\centering
	\includegraphics[width= 0.35\textwidth]{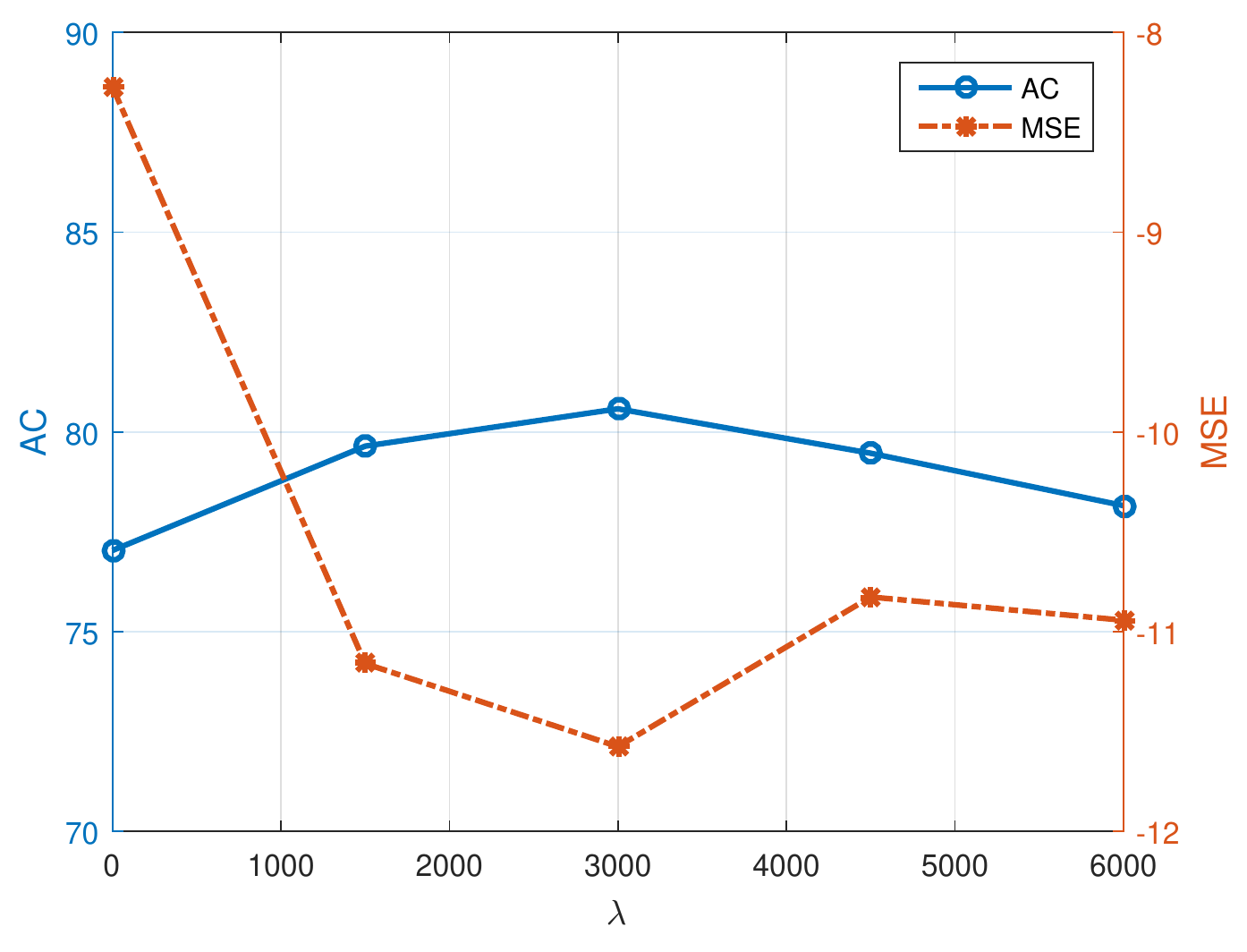}
	\caption{AC and MSE versus parameter $\lambda$}
	\label{fig:param}
\end{figure}

In Fig.~\ref{fig:param}, we show the effect of changing $\lambda$ in the JNKM formulation.
We are particularly interested in this parameter since it plays an essential role in balancing the data fidelity and prior information.
On the other hand, the parameter $\mu$ for enforcing ${\bf Z}$ to be close to ${\bf H}$ can be set to a large number, e.g., 1000,
and $\eta$ for balancing the scaling of the factors can usually be set to a small number -- and the algorithm is not sensitive to these two according to our experience.
Here, the setting is $I = 10, ~J = 100, ~F = 2$, and the number of clusters is $K=2$. The SNRs are set to $\text{SNR}_1 = 5 \text{dB}$ and $\text{SNR}_2 = 30 \text{dB}$.
From Fig.~\ref{fig:param}, we see that both the MSE and AC performance of JNKM is reasonably good
for all the $\lambda$'s under test, although there does exist a certain $\lambda$ giving the best performance ($\lambda=3000$ in this case).

So far we have been working with sparse nonnegative factors.
Let us consider a general $\W$ and a $\H$ with columns in a simplex, i.e. ${\bf 1}^T\H = {\bf 1}^T, ~ \H \geq {\bf 0}$,
which finds various applications in machine learning, e.g., document and hyperspectral pixel clustering / classification \cite{gillis2014robust, gillis2014and}.
We generate $\H$ using the same steps as in the previous simulations with the centroid matrix ${\bf M}(:,k)$'s being generated by putting a cluster near each unit vector, and several centroids randomly;
under such setting, the identifiability conditions of the VolMin model are likely to hold \cite{fu2015blind}.
Note that the entries of ${\bf W}$ are simply drawn from a zero-mean unit-variance i.i.d. Gaussian distribution,
which means that ${\bf W}$ is a dense matrix and the identifiability of NMF does not hold -- which differs from the previous simulations.
{Table~\ref{simu: dense} presents the results.
We see that JNKM works worse relative to the previous simulation, since the generative model is not identifiable via NMF.
However, JVKM works quite well since the VolMin identifiability holds regardless of the structure of ${\bf W}$, so long as it is full column rank. This is also reflected in the MSE performance of VolMin and NMF.}

\begin{table}
	\scriptsize
	\centering	
	\caption{Clustering and factorization accuracy for identifiable VolMin vs. $\text{SNR}_2$, for $I = 50, J = 1000, F = 7, K = 10$.}
	\resizebox{0.49\textwidth}{!}{%
		\begin{tabular}{c|c||c|c|c|c|c|c}
			\cline{1-8}
			\hline
			\hline
			\multicolumn{2}{c||}{$\text{SNR}_2$ [dB]}
			& 3     & 6     & 9     & 12    & 15    & 18   \\
			\hline	
			\hline
			\multicolumn{1}{ c|  }{\multirow{6}{*}{AC[\%]} } & {KM} & 61.01 & 73.82 & 74.59 & 73.53 & 74.66 & 75.04     \\
			\cline{2-8}				
			\multicolumn{1}{ c|  }{} & {RKM} & 62.48 & 75.09 & 78.09 & 76.8  & 74.75 & 76.97     \\
			\cline{2-8}	
			\multicolumn{1}{ c|  }{} & {FKM} & 14.28 & 14.47 & 14.66 & 14.98 & 15    & 15.28      \\
			\cline{2-8}		
			\multicolumn{1}{ c|  }{} & {JNKM } & 58.46 & 75.68 & 85.52 & 89.74 & 91.43 & 92.56       \\
			\cline{2-8}		
			\multicolumn{1}{ c|  }{} & {JVKM } & \textbf{67.9} & \textbf{87.98} & \textbf{95.16} & \textbf{96.54} & \textbf{96.94} & \textbf{97.29}     \\
			\cline{2-8}
			\multicolumn{1}{ c|  }{} & {VolMin-KM } & 64.85 & 81.7  & 87.31 & 89.31 & 88.59 & 89.78    \\ 		
			\hline	
			\multicolumn{1}{ c|  }{\multirow{3}{*}{MSE[dB]} } & {JNKM }&
			-0.11 & 0.03  & 0.22  & 0.21  & 0.34  & 0.62      \\
			\cline{2-8}		
			\multicolumn{1}{ c| }{} & {JVKM } & \textbf{-21.09} & \textbf{-27.75} & \textbf{-29.14} & \textbf{-29.79} & \textbf{-30.7} & \textbf{-30.98}     \\
			\cline{2-8}		
			\multicolumn{1}{ c| }{} & {VolMin-KM } & -11.88 & -15.18 & -15.16 & -15.58 & -15.84 & -15.86    \\
			\hline
			\multicolumn{1}{ c|  }{\multirow{6}{*}{TIME[s]} } &  {KM}  & \textbf{0.14} & \textbf{0.11} & \textbf{0.07} & \textbf{0.06} & \textbf{0.06} & \textbf{0.06}  \\
			\cline{2-8}		
			\multicolumn{1}{ c|  }{} & {RKM} & 0.73  & 0.58  & 0.46  & 0.41  & 0.4   & 0.38  \\
			\cline{2-8}	
			\multicolumn{1}{ c|  }{} & {FKM} & 1.89  & 2.03  & 1.69  & 1.67  & 1.68  & 1.87  \\
			\cline{2-8}	
			\multicolumn{1}{ c|  }{} & {JNKM} & 5.49  & 6.08  & 5.81  & 5.83  & 5.66  & 6.09  \\
			\cline{2-8}	
			\multicolumn{1}{ c|  }{} & {JVKM} & 3.02  & 2.5   & 2.17  & 2.1   & 1.99  & 2.24  \\
			\cline{2-8}	
			\multicolumn{1}{ c|  }{} & {VolMin-KM } & 1.68  & 1.57  & 1.46  & 1.43  & 1.39  & 1.39  \\		
			\cline{1-8}			
		\end{tabular}	
		\label{simu: dense}
	}
\end{table}

In Table~\ref{tab: jntf_km}, we test the joint tensor factorization and latent clustering algorithm (JTKM).
We generate a three-way tensor $\underline{\bf X}\in\mathbb{R}^{I\times J\times L}$ with $I=J=L=30$ and loading factors ${\bf A}\in\mathbb{R}^{I\times F}$, ${\bf B}\in\mathbb{R}^{J\times F}$, ${\bf C}\in\mathbb{R}^{L\times F}$.
To obtain ${\bf A}$ with a cluster structure on its rows, we first generate a centroid matrix ${\bf M}= 2{\bf I} + {\bf 1}{\bf 1}^T$, and then replicate its columns and add noise to create $\tilde{\bf A}$. This way, the rows of $\tilde{\bf A}$ randomly scatter around the rows of ${\bf M}$. Then we let ${\bf A}={\bf D}\tilde{\mathbf{A}}$, where ${\bf D}$ is a diagonal matrix whose diagonal elements are uniformly distributed between zero and one. Here, ${\bf B}$ and ${\bf C}$ are randomly drawn from an i.i.d. uniform distribution between zero and one. Gaussian noise is finally added to the obtained tensor. As in the matrix case, $\text{SNR}_1$ denotes the SNR in the data domain, and $\text{SNR}_2$ the SNR in the latent domain.
In this experiment we set $\text{SNR}_1 = 20 \text{dB}$, $\text{SNR}_2 = 25 \text{dB}$.
As before, to create more severe modeling error so that the situation is more realistic, we finally replace two slabs (i.e., $\underline{\bf X}(:,:,i)$'s)
with elements randomly distributed between zero and one; these slabs mimic outlying data that are commonly seen in practice.

We apply the tensor version of the formulation in \eqref{nmf-reg-vs} to factor the synthesized tensors for various $F=K$.
For each parameter setting, 100 independent trials are performed with randomly generated tensors, and the results are the average of all these trials. As shown in Tab. \ref{tab: jntf_km}, the proposed approach consistently yields higher clustering accuracy, and lower MSEs than plain NTF. This suggests that the clustering regularization does help in better estimating the latent factors, and yields a higher clustering accuracy.

\begin{table}
	\centering
	\caption{MSE and clustering accuracy of JTKM vs. NTF for various $F=K$.}
	\resizebox{.49\textwidth}{!}{%
		\begin{tabular}{c|c|c|c|c|c|c|c|c}
			\hline
			\hline
			\multicolumn{2}{c||}{$F$}&  2     & 3     & 4     & 5     & 6     & 7    & 8   \\
			\hline
			\hline
			\multicolumn{1}{ c  }{\multirow{2}{*}{AC[\%]} } &
			\multicolumn{1}{ |c|| }{JTKM} & \textbf{92.97} & \textbf{74.17} & \textbf{72.33} & \textbf{74.2} & \textbf{76.93} & \textbf{78.4} & \textbf{79.47}   \\
			\cline{2-9}
			\multicolumn{1}{}{}                        &
			\multicolumn{1}{ |c|| }{NTF} & 80.5  & 64.8  & 62    & 62.63 & 62    & 62.2  & 62.7     \\ 			
			\hline
			\multicolumn{1}{ c  }{\multirow{2}{*}{MSE[dB]} } &
			\multicolumn{1}{ |c|| }{JTKM}& \textbf{-22.54} & \textbf{-12.45} & \textbf{-10.04} & \textbf{-10.81} & \textbf{-12.08} & \textbf{-12.92} & \textbf{-13.84}  \\
			\cline{2-9}
			\multicolumn{1}{ c }{}                        &
			\multicolumn{1}{ |c|| }{NTF} &-21.52 & -11.51 & -9.66 & -10.47 & -11.53 & -12.28 & -13.34  \\
			\hline
		\end{tabular}
	}
	\label{tab: jntf_km}
\end{table}

To conclude this section, we present a simulation where the latent data representations lie in different subspaces.
We apply the joint factorization and latent subspace clustering algorithm to deal with this situation.
As a baseline, the latent space sparse subspace clustering (LS3C) method \cite{patel2013latent} is employed.
The idea of LS3C is closely related to FKM, except that the latent clustering part is replaced by sparse subspace clustering.
We construct a dataset with data that lie in two independent two-dimensional subspaces. We set $I = 10$, $J = 200$, $F=4$ and $K=2$;
each subspace contains 100 data columns. As before, we add noise in the latent domain, as well as the data domain. The SNR in data domain is fixed at $\text{SNR}_1 = 30 ~\text{dB}$, and the SNR in the latent domain varies. The parameters of our formulation are set to $\lambda = 1$ and $\mu = 0.5$. For LS3C, we used the code and parameters provided by the authors\footnote{Available online at http://www.rci.rutgers.edu/~vmp93/Software.html}.
The results are shown in Table~\ref{simu: ksub}. As can be seen, our method recovers the factors well, and always gets higher clustering accuracy.

\begin{table}
	\scriptsize
	\centering
	\caption{Simulation comparison of proposed JNKS with LS3C }
	\resizebox{.49\textwidth}{!}{%
	\begin{tabular}{l|c||c|c|c|c|c|c|c}
		\cline{1-8}
		\hline
		\hline
		\multicolumn{2}{c||}{$\text{SNR}_2$ [dB]}
		& 3     & 5     & 7     & 9     & 11    & 13    & 15  \\
		\cline{1-9}	
		\hline
		\hline		
		\multicolumn{1}{ c|  }{\multirow{2}{*}{AC[\%]} } & {LS3C} & 82.9  & 87.45 & 89.51 & 91.9  & 93.74 & 94.14 & 95.81    \\
		\cline{2-9}
		\multicolumn{1}{ c| }{} & JNKS & {\bf 87.65} & {\bf 91.17} & {\bf 92.87} & {\bf 94.39} & {\bf 95.91 }& {\bf 97.15} &{ \bf 97.93}    \\
		\cline{1-9}	
		\multicolumn{1}{ c|  }{\multirow{1}{*}{MSE[dB]} } & {JNKS} & -13.84 & -15.52 & -15.96 & -15.75 & -17.37 & -16.78 & -17.24   \\
		\cline{1-9}				
	\end{tabular}
	\label{simu: ksub}
}
\end{table}

\section{Real-Data Validation}\label{ch: experiment}

\subsection{Document Clustering}
We first test our proposed approach on the document clustering problem.
In this problem, the data matrix ${\bf X}$ is a word-document matrix, the columns of ${\bf W}$ represent $F$ leading topics in this collection of documents, and ${\bf H}(f,\ell)$ denotes the weight that indicates how much document $\ell$ is related to topic $f$. We use a subset of the Reuters-21578 corpus\footnote{Online at: \url{http://www.daviddlewis.com/resources/testcollections/reuters21578/}} as in \cite{cai2011locally}, which contains 8,213 documents from 41 categories. The number of words (features) is 18,933.
Following standard pre-processing, each document is represented as a term-frequency-inverse-document-frequency (tf-idf) vector, and \emph{normalized cut weighting} is applied; see \cite{xu2004document,manning2008introduction} for details.

We apply JNKM and JVKM to the pre-processed data.
A regularized NMF-based approach, namely, \emph{locally consistent concept factorization} (LCCF) \cite{cai2011locally} is employed as the baseline.
LCCF is considered state-of-the-art for clustering the Reuters corpus;
it makes use of data-domain similarity to enforce latent similarity, and it demonstrates superior performance compared to other algorithms on several document clustering tasks. In addition, we also implemented three other methods: spectral clustering (SC) \cite{ng2002spectral}, \cite{von2007tutorial} and NMF followed by $K$-means (NMF-KM)\cite{xu2003document}, and RKM \cite{de1994k}.
FKM \cite{vichi2001factorial} is not applied here since this method is not scalable: each iteration of FKM requires taking an eigenvalue decomposition on a large matrix (in our case a $18933\times 18933$ matrix).

\begin{figure}[h!]
	\centering
	\includegraphics[width= 0.49\textwidth]{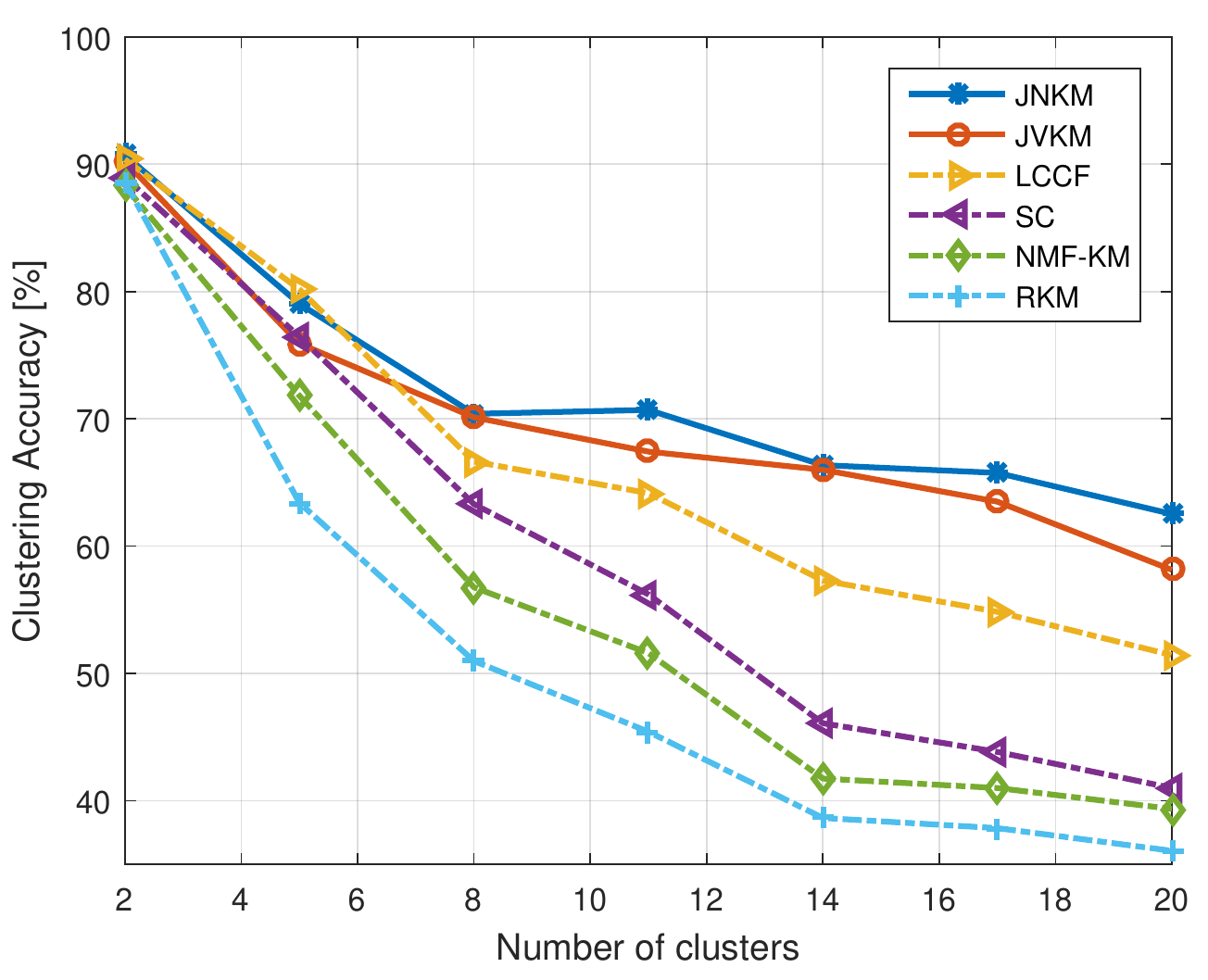}
	\caption{Clustering accuracy with different number of clusters on Reuters-21578 dataset}
	\label{fig:text}
\end{figure}

Fig. \ref{fig:text} presents our experiment results. We test the above mentioned methods on the Reuters-21578 data for various $K$ and use $F=K$ for the methods that perform factorization.
For each $K$, we perform 50 Monte-Carlo trials by randomly selecting $K$ clusters out of the total 41 clusters (categories), and report the performance by comparing the results with the ground truth. Clustering performance is measured by clustering accuracy.
The averaged result shows that our proposed methods, i.e., JNKM and JVKM, outperform the other methods under test.
For simpler clustering tasks, e.g., when the number of clusters is small, the difference in clustering accuracy is relatively small. However, when $K$ becomes larger, the proposed methods get much higher accuracy than the others. For example, for $K =  20$, our JNKM formulation improves clustering accuracy by $11.1\%$ compare to LCCF.
{For all the $K$'s under test, JVKM performs slightly worse than JNKM in terms of accuracy,
but it has a simpler update strategy and thus is faster.}

\subsection{Image Clustering }\label{ch:mnist}
We also test the proposed approach using image data.
Image data is known to be suitable for subspace clustering,
and thus we apply the JNKS algorithm here.
We compare our method with state-of-the-art subspace clustering methods, namely, the sparse subspace clustering (SSC) \cite{elhamifar2013sparse} and LS3C \cite{patel2013latent}.
We evaluate these methods on the widely used MNIST \footnote{Online available: http://yann.lecun.com/exdb/mnist/} dataset.
The MNIST dataset contains images of handwritten digits, from $0$ to $9$. We use only the testing subset of the dataset, which contains $10000$ images, with each digit having $\approx$ $1000$ images. Each $28\times 28$ gray-level image is vectorized into a $784 \times 1$ vector. It was pointed out in \cite{hastie1998metrics} that vectors of each digit lie approximately in a 12-dimensional subspace \eqref{eq:NMF_Ksub} and we adopt this value in our experiments.

\begin{figure}[h!]
	\centering
	\includegraphics[width= 0.5\textwidth]{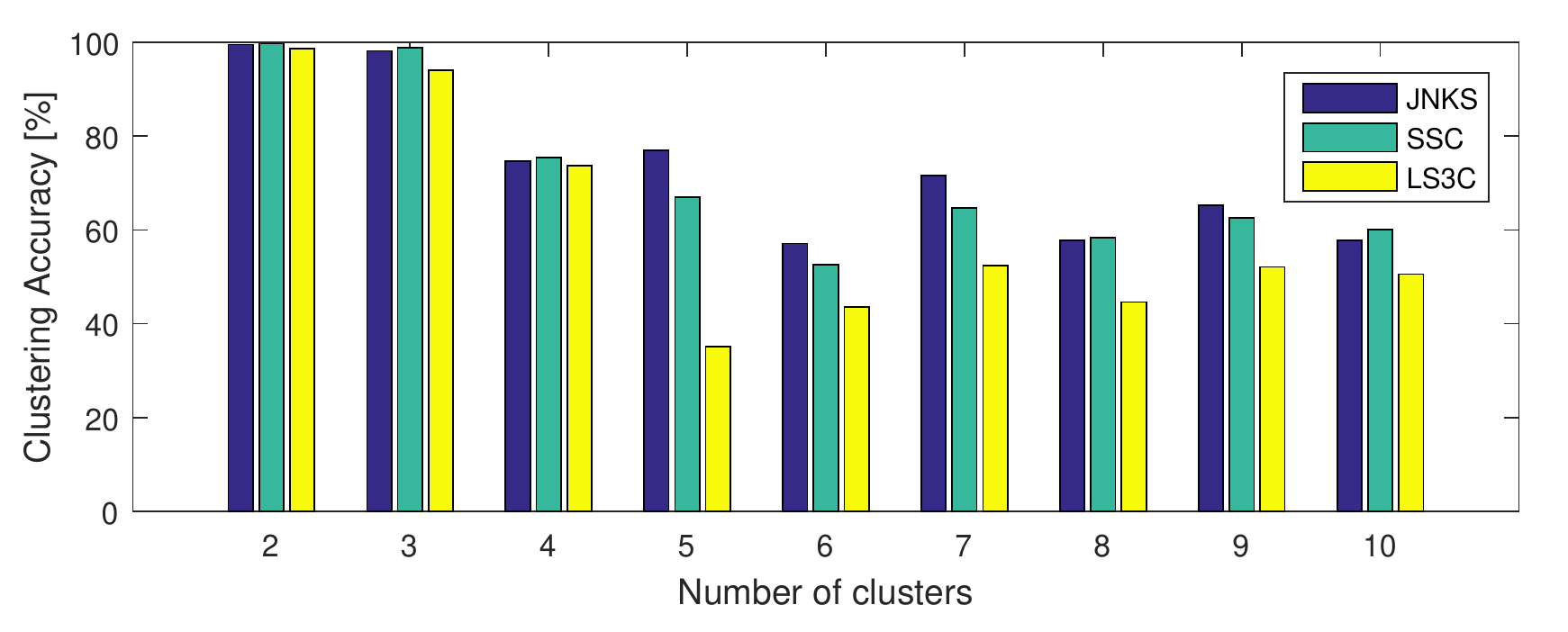}
	\caption{Clustering accuracy with different number of clusters on MNIST dataset, each cluster has 200 samples.}
	\label{fig:ksub_f}
\end{figure}

Fig.~\ref{fig:ksub_f} shows how clustering accuracy changes with the number of clusters.
For each number of clusters, we perform $50$ trials, each time randomly picking digits to perform clustering. In each trial, we randomly pick $200$ images for each digit. For example, for 2 clusters, each trial we will have in total $400$ images. Note that JNKS outperforms SSC and LS3C when the number of cluster is moderate (5-7), and remains competitive with SSC, LS3C in all other cases. 

\begin{figure}[h!]
	\centering
	\includegraphics[width= 0.5\textwidth]{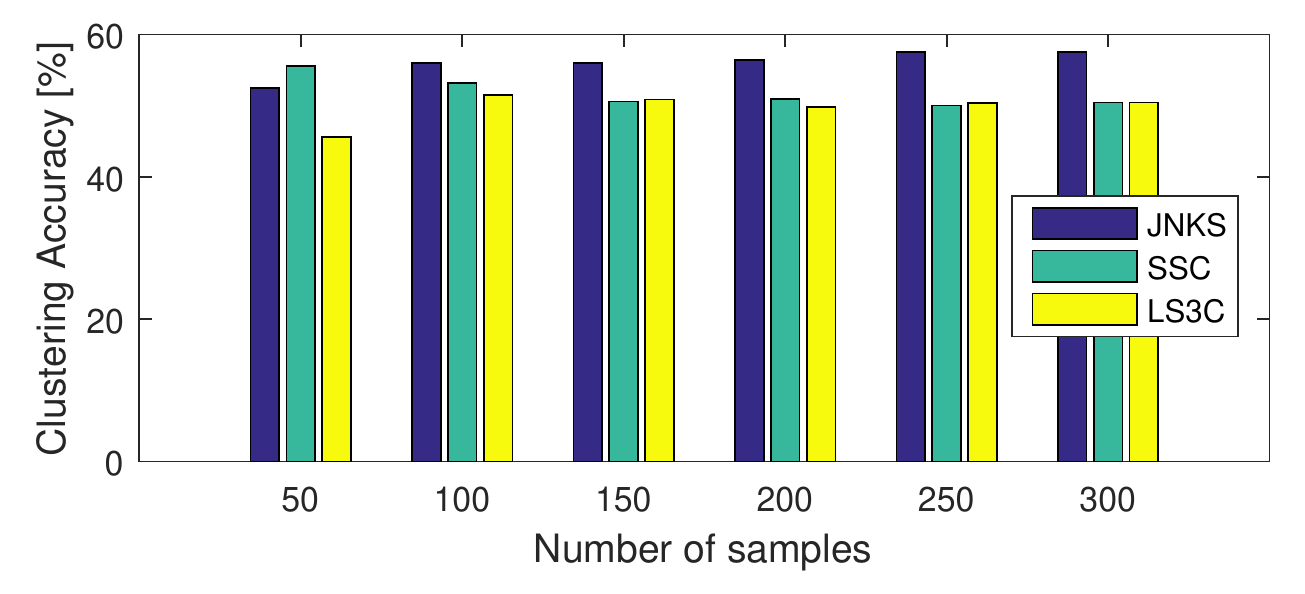}
	\caption{Clustering accuracy with different number of samples, all the 10 digits (clusters) from MNIST dataset.}
	\label{fig:ksub_samples}
\end{figure}

Fig.~\ref{fig:ksub_samples} shows the results with all the 10 clusters under various number of samples of each cluster. We also average the results over 50 random trials as before.
As can be seen, when the number of samples is small, JNKS and SSC
get similar performance. With more data samples per cluster, however, the proposed method gets consistently higher accuracy.

\subsection{Tensor Social Network Data Analysis}
In this subsection, we apply the proposed JTKM algorithm to analyze the Enron email dataset, made public by the U.S. Department of Justice, as part of the Enron investigation. The data that we used contain communication counts between $184$ employees over a period of $44$ months, arranged in a three-way tensor of size $184\times 184\times 44$. The $(i,j,l)$ entry of this tensor indicates the number of emails employee $i$ sent to employee $j$ in month $l$ \footnote{Online available at \url{http://cis.jhu.edu/~parky/Enron/enron.html}}.
This dataset was analyzed with low-rank tensor factorization models \cite{bader2006temporal}, \cite{papalexakis2013k}.

Enron filed for bankruptcy in late 2001, corresponding to months 36 $\sim$ 38 out of 44. Thus the 44 months can be roughly divided into pre-crisis, crisis, and post-crisis period.
Therefore, we impose a cluster structure on the time-mode factor and come up with the following formulation:
\begin{align}\label{ntf-l1}
\min_{\begin{subarray}{c}{\bf A},{\bf B},{\bf C}\\{\bf S},{\bf M}.\end{subarray}} & ~~\left\| \X_{(1)} -({\bf C}\odot{\bf B}){\bf A}^T\right\|^2_F +\eta\left(\left\| \mathbf{A}\right\|_1+ \|{\bf B}\|_1\right)\nonumber\\
&~~ + \lambda\|{\bf C} - \S\M\|_F^2\nonumber\\
{\rm s.t.} &~ \mathbf{A},{\bf B},{\bf C}\geq {\bf 0},\nonumber\\
&~\|{\bf S}(i,:)\|_0=1,~{\bf S}(i,k)\in\{0,1\},~\forall i,k,
\end{align}
where we set $K=3$ and ${\bf C}$ denotes the time-mode factor.
Following \cite{papalexakis2013k}, we also use $\ell_1$-norm regularizers on ${\bf A}$ and ${\bf B}$, to control scaling but also promote sparsity at the same time. We compare our latent clustering formulation with multi-way co-clustering \cite{papalexakis2013k} and plain NTF. As suggested by previous works on this dataset, we aim at identifying 5 groups of people, so we set $F = 5$ for all methods. Other parameters in formulation \eqref{ntf-l1} are set to $\lambda = 500$, $\eta = 5$. The dataset is pre-processed as suggested in \cite{bader2006temporal}, i.e., the nonzero values are transformed using $x' = \text{log}_2x+1$ to compress the dynamic range. After getting the results, we derive the clustering result from the estimated $\A$ factor. To measure the clustering quality, we first remove from the result the $71$ employees who do not have clear roles, usually temporary employees and interns. The remaining $113$ people have one or more of the four roles: legal (e.g., lawyers), executives (e.g. VPs, CEOs), trading, and pipeline operations.

We obtain qualitatively consistent cluster structure as reported in previous works. The `Legal' cluster identified by the three methods is tabulated in Table \ref{tab:enron}. For this cluster, the proposed method gets the same result as the sparse co-clustering method \cite{papalexakis2013k}, both of which are much cleaner than the result of NTF. In total, the proposed method gets 19 mis-classified employees compared to 21 for the sparse co-clustering \cite{papalexakis2013k} and 24 for plain NTF, respectively.
\begin{table*}[t]
			\centering
			\caption{The Legal cluster identified by the three methods, Bold entries are miss-clustered.}
			\resizebox{\textwidth}{!}{
			\begin{tabular}{l|l|l}
				\hline
				Proposed & Co-clustering & NTF \\
				\hline
				Debra Perlingiere, Legal Specialist ENA Legal & Debra Perlingiere, Legal Specialist ENA Legal & Debra Perlingiere, Legal Specialist ENA Legal \\
				Elizabeth Sager,VP and Asst Legal Counsel ENA Legal & Elizabeth Sager,VP and Asst Legal Counsel ENA Legal & Elizabeth Sager, VP and Asst Legal Counsel ENA Legal \\
				\textbf{Jason Williams ,Trader Central Desk Gas Trading} & \textbf{Jason Williams ,Trader Central Desk Gas Trading} & \textbf{Eric Saibi, Trader } \\
				Jeff Hodge, Asst General Counsel ENA Legal & Jeff Hodge, Asst General Counsel ENA Legal & Mark Taylor, Manager Financial Trading Group ENA Legal \\
				Kay Mann , Lawyer  & Kay Mann , Lawyer  & \textbf{Jason Williams, Trader Central Desk Gas Trading} \\
				\textbf{Kim Ward, Manager West Gas Origination} & \textbf{Kim Ward, Manager West Gas Origination} & Jeff Hodge, Asst General Counsel ENA Legal \\
				Marie Heard,Senior Legal Specialist ENA Legal & Marie Heard,Senior Legal Specialist ENA Legal & Kay Mann, Lawyer  \\
				Mark Haedicke, Managing Director ENA Legal & Mark Haedicke, Managing Director ENA Legal & \textbf{Kevin Ruscitti, Trader Central Desk Gas Trading} \\
				Mark Taylor, Manager Financial Trading Group ENA Legal & Mark Taylor, Manager Financial Trading Group ENA Legal & \textbf{Kim Ward, Manager West Gas Origination} \\
				Sara Shackleton, Employee ENA Legal & Sara Shackleton, Employee ENA Legal & Marie Heard, Senior Legal Specialist ENA Legal \\
				Stacy Dickson, Employee ENA Legal & Stacy Dickson, Employee ENA Legal & Mark Haedicke, Managing Director ENA Legal \\
				Stephanie Panus , Senior Legal Specialist ENA Legal & Stephanie Panus , Senior Legal Specialist ENA Legal & Mark Haedicke, Managing Director ENA Legal \\
				Susan Bailey, Legal Assistant ENA Legal & Susan Bailey, Legal Assistant ENA Legal & Mark Taylor, Manager Financial Trading Group ENA Legal \\
				Tana Jones , Employee Financial Trading Group ENA Legal & Tana Jones , Employee Financial Trading Group ENA Legal & \textbf{Michelle Lokay , Admin. Asst. Transwestern Commercial Group} \\
				&       & Sara Shackleton , Employee ENA Legal \\
				&       & Stacy Dickson, Employee ENA Legal \\
				&       & Stephanie Panus, Senior Legal Specialist ENA Legal \\
				&       & \textbf{Steven South,Director West Desk Gas Trading} \\
				&       & Susan Bailey, Legal Assistant ENA Legal \\
				&       & \textbf{Kim Ward, Manager West Gas Origination} \\
				&       & Tana Jones, Employee Financial Trading Group ENA Legal \\
				\hline
			\end{tabular}%
		}
		\label{tab:enron}%
\end{table*}%

\section{Conclusions}
We proposed a framework for joint factorization and latent clustering, motivated by the fact that many datasets exhibit better
cluster structure in {\em some} reduced-dimension domain. Our proposed framework leverages the identifiability of certain matrix and tensor factorization models together with a latent clustering prior to produce more discriminative latent representations that are suitable for clustering, and more accurate latent factors for estimation purposes. Carefully designed optimization objectives were proposed for joint factorization and $K$-means/$K$-subspace clustering.
Alternating optimization-type algorithms were proposed for handling the proposed formulations, and judicious  
simulations as well as experiments with benchmark document, image, and social network data showed that the proposed approaches offer promising performance, and can outperform state-of-art methods for the respective datasets.  

\section*{Acknowledgment}
The authors would like to thank Prof. Eva Ceulemans for pointing out relevant references on RKM and FKM.

\bibliography{ref}{}

\end{document}